\documentclass[letterpaper, 10 pt, conference]{ieeeconf}  

\IEEEoverridecommandlockouts                              

\usepackage{graphicx}                   
\usepackage{mathptmx}                   
\usepackage{times}                      
\usepackage{amsmath}                    
\usepackage{amssymb}                    
\usepackage{algorithm}
\usepackage{algorithmic}
\usepackage{subfig}
\usepackage{balance}
\graphicspath{{figures/}}

\newcommand{\comment}[1]{#1}

\title{\LARGE \bf
Meta Inverse Reinforcement Learning via Maximum Reward Sharing for Human Motion Analysis}
\author{Kun Li$^{1}$, Joel W. Burdick$^{1}$
\thanks{*This work was
supported by the National Institutes of Health, NIBIB.} \thanks{$^{1}$Kun Li and Joel W. Burdick are
with Department of Mechanical and Civil Engineering, California Institute of Technology, Pasadena,
CA 91125, USA {\tt\small kunli@caltech.edu}
}%
}
\begin{document}
\maketitle
\thispagestyle{empty}
\pagestyle{empty}
\begin{abstract}
  This work handles the inverse reinforcement learning (IRL) problem where only a small number of
  demonstrations are available from a demonstrator for each high-dimensional task, insufficient to
  estimate an accurate reward function. Observing that each demonstrator has an inherent reward for
  each state and the task-specific behaviors mainly depend on a small number of key states, we
  propose a meta IRL algorithm that first models the reward function for each task as a distribution
  conditioned on a baseline reward function shared by all tasks and dependent only on the
  demonstrator, and then finds the most likely reward function in the distribution that explains the
  task-specific behaviors. We test the method in a simulated environment on path planning tasks with
  limited demonstrations, and show that the accuracy of the learned reward function is significantly
  improved. We also apply the method to analyze the motion of a patient under rehabilitation.
\end{abstract}
\section{Introduction}
\label{irl::intro}
Inverse reinforcement learning (IRL) \cite{irl::irl1} algorithms estimate a reward function that
explains the motions demonstrated by an operator or other agents on a task described by a Markov
Decision Process (MDP) \cite{irl::rl}. The recovered reward function can be used by a robot to
replicate the demonstrated task \cite{irl::irl2}, or by an algorithm to analyze the demonstrator's
preference \cite{irl::motionanalysis}. Therefore, IRL algorithms can make multi-task robot control
simpler by alleviating the need to explicitly set a cost function for each task, and make robot
friendlier by personalizing services based on the recovered condition and preference of the
operator. 

The accuracy of the recovered function depends heavily on the ratio of visited states in the
demonstrations to the whole state space, because the demonstrator's motion policy can be estimated
more accurately if every state is repeatedly visited. However, the ratio is low for many useful
applications, since they usually have huge or high-dimensional state spaces, while the
demonstrations are relatively rare for each task. For example, in a path planning task on a mild
$100*100$ grid, the demonstrator chooses paths based on the destination, but may not move to the
same destination hundreds of times in practice. For robot manipulation tasks based on ordinary
$640*480*3$ RGB images, east task specifies a final result, but it is expensive to repeat each task
millions of times. For human motion analysis, it is physically improbable to follow an instruction
thousands of times in the huge state space of human poses. Therefore, it is difficult to estimate an
accurate reward function for a single task with limited data.

In practice, usually multiple tasks can be observed from the same demonstrator, and the problem of
rare demonstrations can be handled by combining data from all tasks, hence the meta-learning
problem. Existing solutions mainly classification problems, like using the data from all tasks to
learn an optimizer for each task, using the data from all tasks to learn a metric space where a
single task can be more accurate with limited data, using the data from all tasks to learn a good
initialization or a good initial parameter for each task, etc. Some of these methods are applicable
to inverse reinforcement learning problems, but they mainly consider transfer of motion policy. 

\begin{figure}
  \subfloat[]{\includegraphics[width=0.2\textwidth]{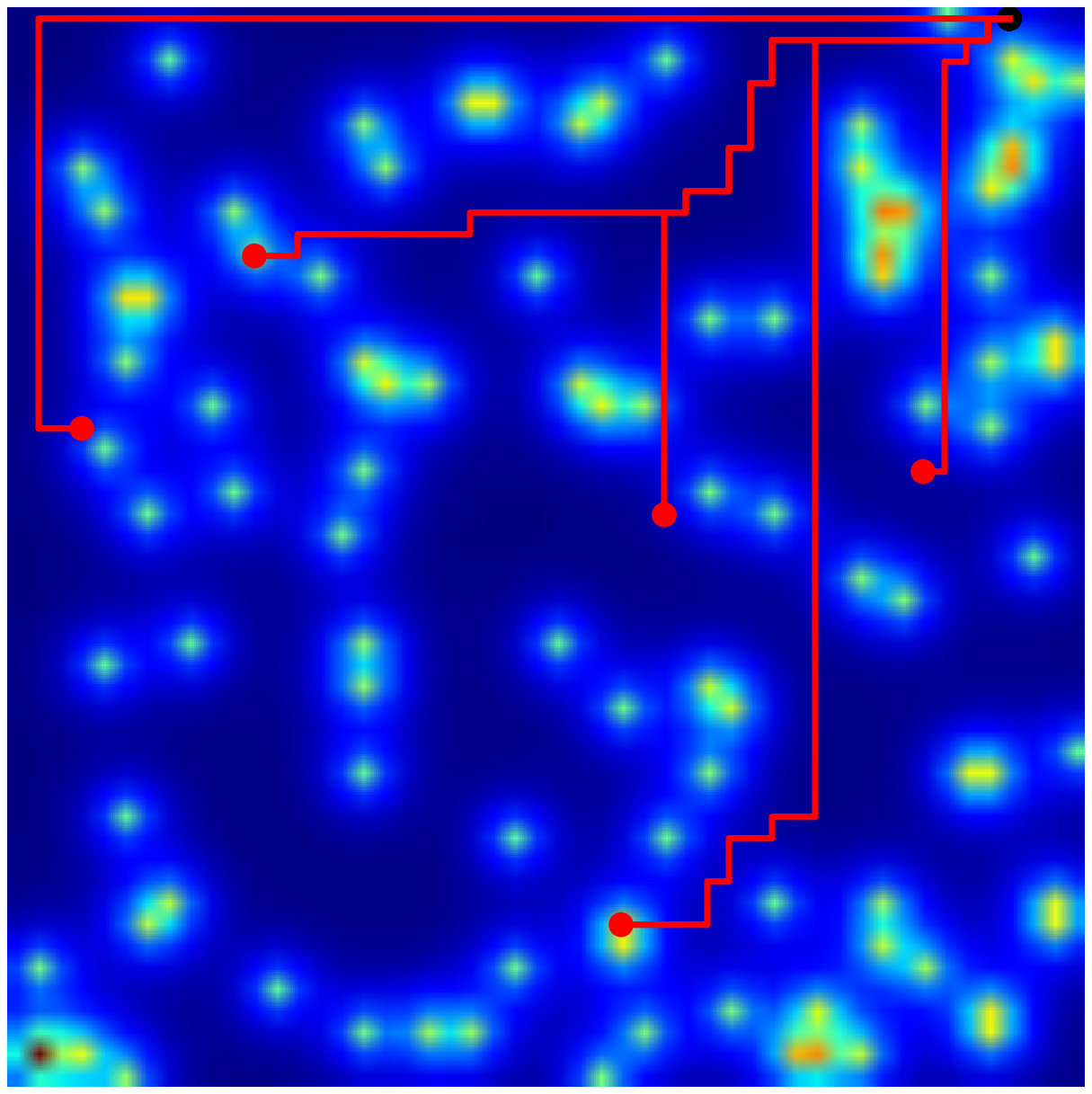}\label{fig:path1}}
  \subfloat[]{\includegraphics[width=0.2\textwidth]{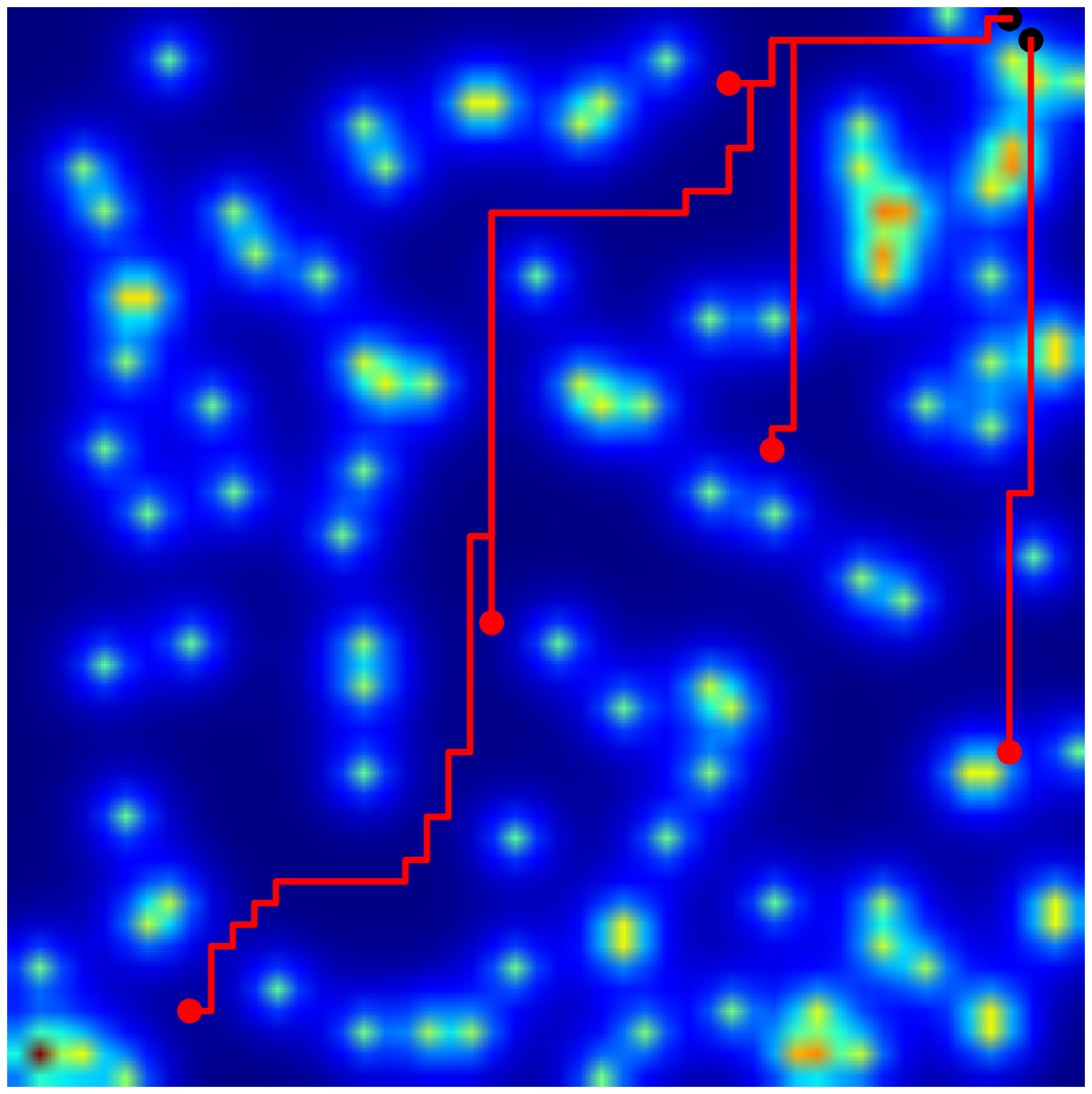}\label{fig:path2}}

  \subfloat[]{\includegraphics[width=0.2\textwidth]{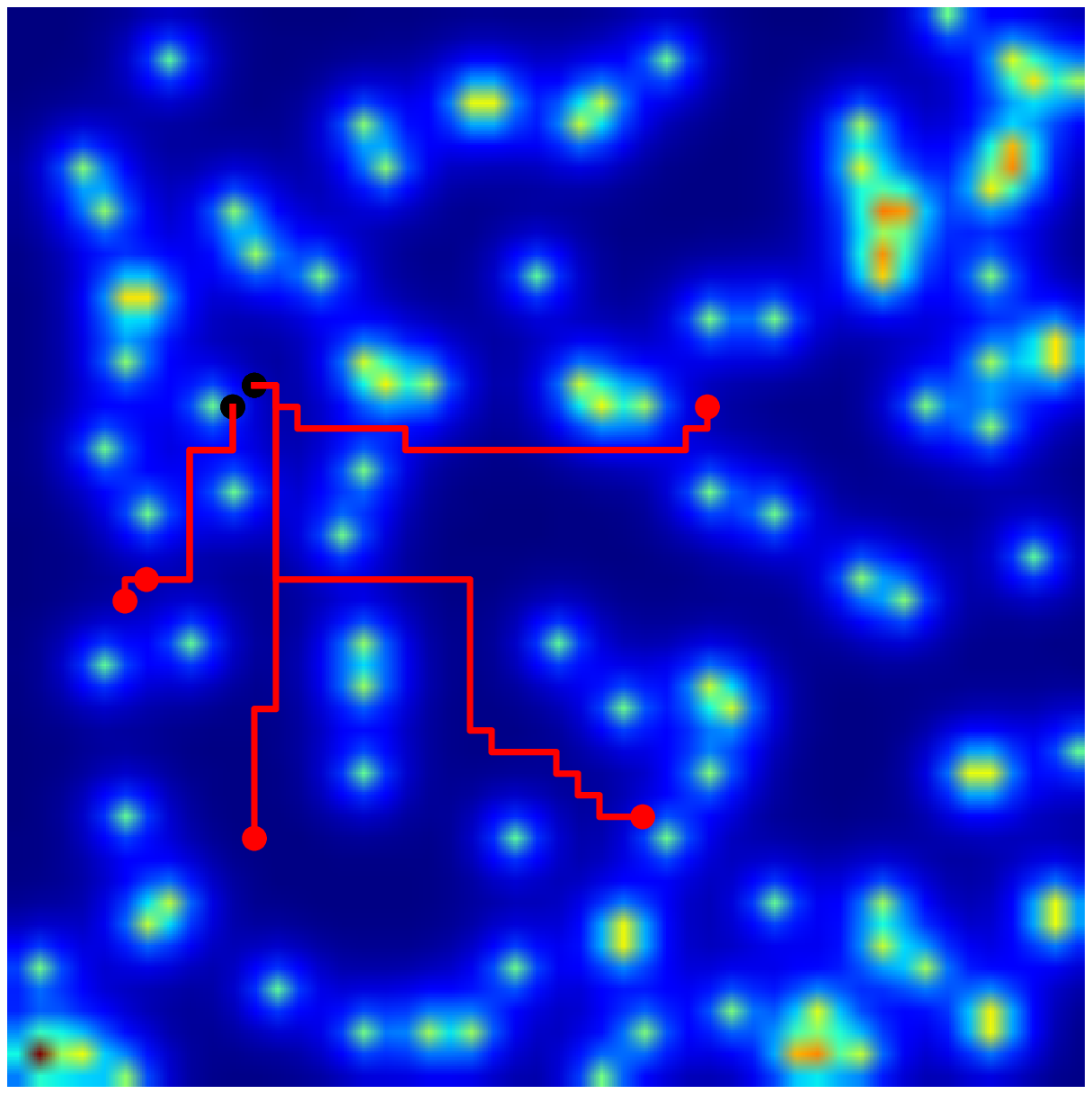}\label{fig:path3}}
  \subfloat[]{\includegraphics[width=0.2\textwidth]{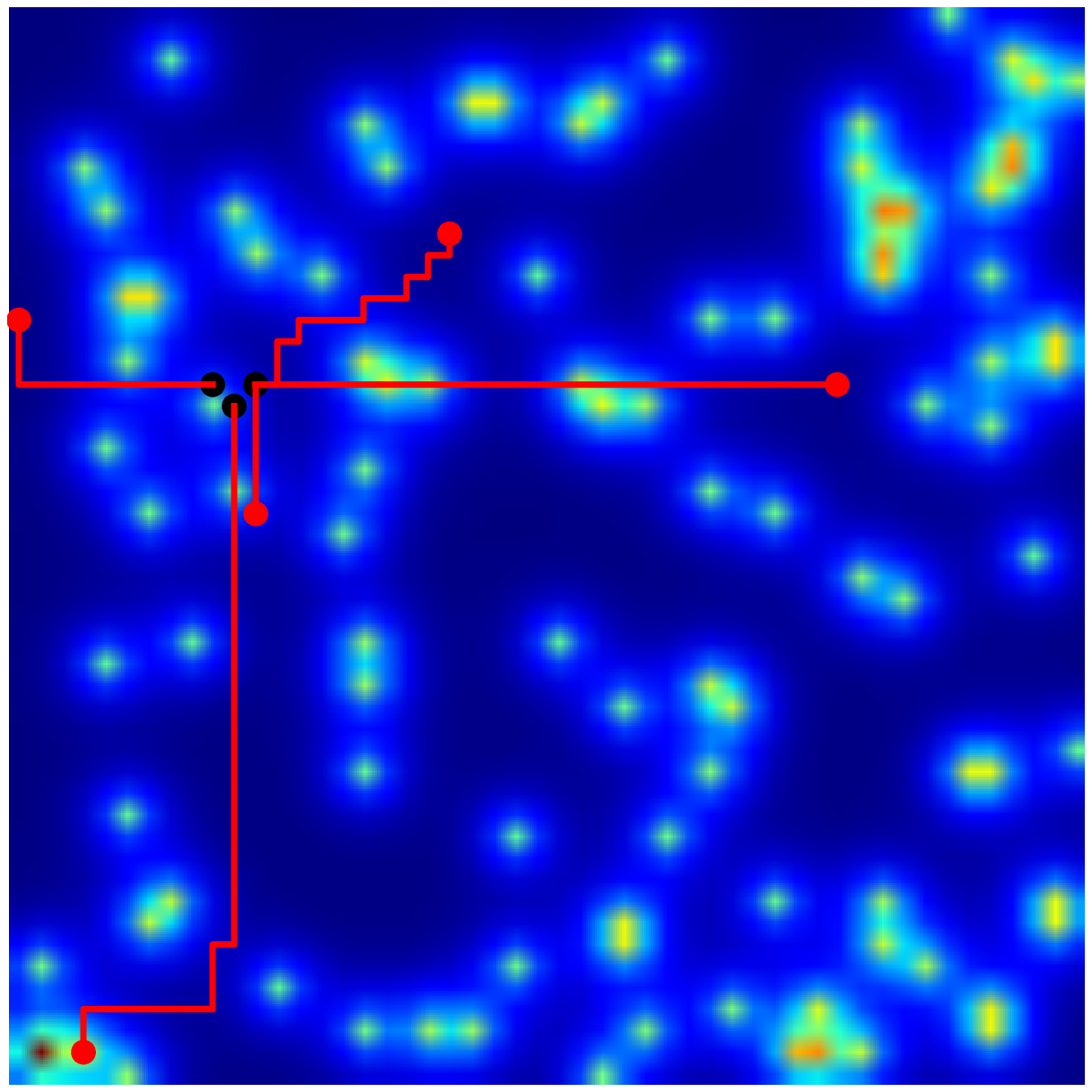}\label{fig:path4}}
  \caption{Different behaviors under different goal states and goal rewards: Figure \ref{fig:path1}
  and Figure \ref{fig:path2} share the same goal state, but the goal reward of Figure
  \ref{fig:path2} is larger than Figure \ref{fig:path1}.Figure \ref{fig:path3}
  and Figure \ref{fig:path4} share the same goal state, but the goal reward of Figure
  \ref{fig:path4} is larger than Figure \ref{fig:path3}. Five trajectories are plotted in each
  figure, where red dots denote the starting point and black dots denote the ending point.}
  \label{fig:examplepath}
\end{figure}

In many IRL applications, we observe that a demonstrator usually has an inherent reward for each
state, materialized as the innate state preferences of a human, the hardware-dependent cost function
of a robot, the default structure of an environment, etc. For a given task, the demonstrators are
usually reluctant to drastically change the inherent reward function to complete the task; instead,
they alter the innate reward function minimally to generate a task-specific reward function and plan
the motion. For example, in path planning, the C-space of a mobile robot at home rarely changes, and
the robot's motion depends on the goal state; in human motion analysis, the costs of different
poses are mostly invariant, while the actual motion depends on the desired directions.
 
Based on this observation, we propose a meta inverse reinforcement learning algorithm by maximizing
the shared rewards among all tasks. We model the reward function for each task as a probabilistic
distribution conditioned on an inherent baseline function, and estimate the most likely reward
function in the distribution that explains the observed task-specific demonstrations. 

We review existing IRL and meta-learning algorithms in Section \ref{irl::related}, and then
introduce the proposed method in Section \ref{irl::metairl}. We show a simulated experiments for
evaluation and a real-world experiment for application in Section \ref{irl::experiments}, with
conclusions in Section \ref{irl::conclusions}.

\section{Related Works}
\label{irl::related}
The idea of inverse optimal control is proposed by Kalman \cite{irl::kalman}, white the inverse
reinforcement learning problem is firstly formulated in \cite{irl::irl1}, where the agent observes
the states resulting from an assumingly optimal policy, and tries to learn a reward function that
makes the policy better than all alternatives. Since the goal can be achieved by multiple reward
functions, this paper tries to find one that maximizes the difference between the observed policy
and the second best policy. This idea is extended by \cite{irl::maxmargin}, in the name of
max-margin learning for inverse optimal control. Another extension is proposed in \cite{irl::irl2},
where the purpose is not to recover the real reward function, but to find a reward function that
leads to a policy equivalent to the observed one, measured by the amount of rewards collected by
following that policy.

Since a motion policy may be difficult to estimate from observations, a behavior-based method is
proposed in \cite{irl::maxentropy}, which models the distribution of behaviors as a maximum-entropy
model on the amount of reward collected from each behavior. This model has many applications and
extensions. For example, \cite{irl::sequence} considers a sequence of changing reward functions
instead of a single reward function. \cite{irl::gaussianirl} and \cite{irl::guidedirl} consider
complex reward functions, instead of linear one, and use Gaussian process and neural networks,
respectively, to model the reward function. \cite{irl::pomdp} considers complex environments,
instead of a well-observed Markov Decision Process, and combines partially observed Markov Decision
Process with reward learning. \cite{irl::localirl} models the behaviors based on the local
optimality of a behavior, instead of the summation of rewards.  \cite{irl::deepirl} uses a
multi-layer neural network to represent nonlinear reward functions.

Another method is proposed in \cite{irl::bayirl}, which models the probability of a behavior as the
product of each state-action's probability, and learns the reward function via maximum a posteriori
estimation. However, due to the complex relation between the reward function and the behavior
distribution, the author uses computationally expensive Monte-Carlo methods to sample the
distribution. This work is extended by \cite{irl::subgradient}, which uses sub-gradient methods to
simplify the problem.  Another  extensions is shown in \cite{irl::bayioc}, which tries to find a
reward function that matches the observed behavior. For motions involving multiple tasks and varying
reward functions, methods are developed in \cite{irl::multirl1} and \cite{irl::multirl2}, which try
to learn multiple reward functions. 

Most of these methods need to solve a reinforcement learning problem in each step of reward
learning, thus practical large-scale application is computationally infeasible. Several methods are
applicable to large-scale applications. The method in \cite{irl::irl1} uses a linear approximation
of the value function, but it requires a set of manually defined basis functions. The methods in
\cite{irl::guidedirl,irl::relative} update the reward function parameter by minimizing the relative
entropy between the observed trajectories and a set of sampled trajectories based on the reward
function, but they require a set of manually segmented trajectories of human motion, where the
choice of trajectory length will affect the result. Besides, these methods solve large-scale
problems by approximating the Bellman Optimality Equation, thus the learned reward function and Q
function are only approximately optimal. In our previous work, we proposed an approximation method
that guarantees the optimality of the learned functions as well as the scalability to large state
space problems \cite{irl::fairl}.

To learn a model from limited data, meta learning algorithms are developed. A survey of different
work is given in \cite{irl::metasurveyclassification}, viewing meta-learner as a way to improve
biases for base-learners. The method in \cite{irl::metamemory} uses neural memory machine to do the
meta learning.  The method in \cite{irl::representation} minimizes the representations. The method
in \cite{irl::learngradient} learns by gradient descent. The method \cite{irl::hypernetworks} learns
optimizers.

Meta learning algorithms are also applied to reinforcement learning problems. The method in
\cite{irl::metarl} tunes meta parameters for reinforcement learning, learning rate for TD learning,
action selection trade-off, and discount factor.  The method in \cite{irl::metamultirl} uses one
network to play multiple games. The method in \cite{irl::rl2} trains reinforcement learning with
slower rl. The method in \cite{irl::maml} learns a good initial parameter that reaches optimal
parameters with limited gradient descent.  Meta learning in inverse reinforcement learning focuses
on imitation learning, like one-shot imitation learning\cite{irl::osil}.

\section{Meta Inverse Reinforcement Learning}
\label{irl::metairl}
\subsection{Meta Inverse Reinforcement Learning}
We assume that an agent needs to handle multiple tasks in an environment, denoted by
$\{\mathbb{T}_i|i=1,N_\mathbb{T}\}$, where $\mathbb{T}_i$ denotes the $i_{th}$ task and
$N_\mathbb{T}$ denotes the number of tasks.

We describe a task $\mathbb{T}_i$ as a Markov Decision Process, consisting of the following
variables:
\begin{itemize}
  \item $S_i=\{s\}$, a set of states
  \item $A_i=\{a\}$, a set of actions
  \item $P_{i,ss'}^a$, a state transition function that defines the probability that state $s$
    becomes $s'$ after action $a$.
  \item $R_i=\{r_i(s)\}$, a reward function that defines the immediate reward of state $s$.
  \item $\gamma_i$, a discount factor that ensures the convergence of the MDP over an infinite
    horizon.
\end{itemize}

For a task $\mathbb{T}_i$, the agent performs a set of demonstrations
$\zeta_i=\{\zeta_{i,j}|j=1,\cdots,N_{\zeta_i}\}$, represented by $N_{\zeta_i}$ sequences of
state-action pairs:
\[\zeta_{i,j}=\{(s^t_{i,j},a^t_{i,j})|t=0,\cdots,N_{\zeta_{i,j}}\},\]
where $N_{\zeta_{i,j}}$ denotes the length of the $j_{th}$ sequence $\zeta_{i,j}$. Given the
observed sequences $\zeta=\{\zeta_i|i=1,\cdots,N_\mathbb{T}\}$ for the $N_\mathbb{T}$ tasks, inverse
reinforcement learning algorithms try to recover a reward function $r_i(s)$ for each task.


Our key observation in multi-task IRL is that the demonstrator has an inherent reward function
$r_b(s)$, generating a baseline reward for each state in all tasks. To complete the $i_{th}$ task,
the agent generates a reward function $r_i(s)$ from a distribution $P(r_i|r_b)$ conditioned on
$r_b(s)$ to plan the motion. Therefore, the motion $\zeta_i$  is generated as:
\[P(\zeta_i|r_i)P(r_i|r_b)\]

For the $i_{th}$ task, we want to find the most likely $r_i(s)$ sampled from $P(r_i|r_b)$ that
explains the demonstration $\zeta_i$. Assuming all the tasks are independent from each other, the
following joint distribution is formulated:
\[\prod_{i=1}^{N_{\mathbb{T}}}P(\zeta_i|r_i)P(r_i|r_b) \]

The reward functions can be found via maximum-likelihood estimation:
\begin{align}
  \min_{r_b(s),r_1(s),\cdots,r_{\mathbb{T}}(s)\in \mathbb{F}} \sum_{i=1}^{N_{\mathbb{T}}} \underbrace{L_i(\zeta_i,r_i(s))}_{\text{IRL
  loss}}+\underbrace{L(r_i(s),r_b(s))}_{\text{reward sharing loss}}
  \label{equation:metaoptimization}
\end{align}
where $\mathbb{F}$ denotes a function space, $L_i(\zeta_i,r_i(s))$ is the negative loglikelihood of
$P(\zeta_i|r_i)$, and $L(r_i(s),r_b(s))$ is the negative loglikelihood $P(r_i|r_b)$.

\subsection{Loss for Inverse Reinforcement Learning}
While many solutions exist for the inverse reinforcement learning problem, we adopt the solution
based on function approximation developed in our earlier work \cite{irl::fairl} to handle the
practical high-dimensional state spaces.

The core idea of the method is to approximate the Bellman Optimality Equation \cite{irl::rl} with a
function approximation framework. The Bellman Optimality Equation is given as:
\begin{align}
  &V^*_i(s)=\max_{a\in A}\sum_{s'|s,a}P_{i,ss'}^a[r_i(s')+\gamma*V_i^*(s')],\\
  &Q^*_i(s,a)=\sum_{s'|s,a}P_{i,ss'}^a[r_i(s')+\gamma*\max_{a'\in A}Q_i^*(s',a')].
\end{align}
It is computationally prohibitive to solve in high-dimensional state spaces.

But with a parameterized \textit{VR function}, we describe the summation of the reward function and
the discounted optimal value function as:
\begin{equation}
  f_i(s,\theta_i)=r_i(s)+\gamma*V_i^*(s),
  \label{equation:approxrewardvalue}
\end{equation}
where $\theta_i$ denotes the parameter of \textit{VR function}. The function value of a state is named
as \textit{VR value}.

Substituting Equation \eqref{equation:approxrewardvalue} into Bellman Optimality Equation, the
optimal Q function is given as:
\begin{equation}
  Q_i^*(s,a)=\sum_{s'|s,a}P_{i,ss'}^af_i(s',\theta_i),
  \label{equation:approxQ}
\end{equation}
the optimal value function is given as:
\begin{align}
  V_i^*(s)=\max_{a\in A}\sum_{s'|s,a}P_{i,ss'}^af_i(s',\theta_i),
  \label{equation:approxV}
\end{align}
and the reward function can be computed as:
\begin{align}
  r_i(s)=f_i(s,\theta_i)-\gamma*\max_{a\in A}\sum_{s'|s,a}P_{i,ss'}^af_i(s',\theta_i).
  \label{equation:approxR}
\end{align}

This framework avoids solving the Bellman Optimality Equation. Besides, this formulation can
be generalized to other extensions of Bellman Optimality Equation by replacing the $max$ operator
with other types of Bellman backup operators. For example, $V^*(s)=\log \sum_{a\in A}\exp Q^*(s,a)$ is
used in the maximum-entropy method\cite{irl::maxentropy}; $V^*(s)=\frac{1}{k}\log \sum_{a\in A}\exp
k*Q^*(s,a)$ is used in Bellman Gradient Iteration \cite{irl::BGI}.  

To apply this framework to IRL problems, this work chooses a motion model $p(a|s)$ based on the
optimal Q function $Q_i^*(s,a)$ \cite{irl::bayirl}:
\begin{equation}
  P(a|s)=\frac{\exp{b*Q_i^*(s,a)}}{\sum_{\tilde{a}\in
  A}\exp{b*Q_i^*(s,\tilde{a})}},
  \label{equation:motionmodel}
\end{equation} 
where $b$ is a parameter controlling the degree of confidence in the agent's ability to choose
actions based on Q values. Other models can also be used, like $p(a|s)=\exp(Q(s,a)-V(s))$ in
\cite{irl::maxentropy}.

Assuming the approximation function is a neural network, the parameter $\theta_i=\{w,b\}$-weights and
biases, the negative log-likelihood of $P(\zeta_i|\theta_i)$ is given by:
\begin{align}
  L_i(\theta_i)=-\sum_{(s,a)\in\zeta_i}(b*Q_i^*(s,a)-\log{\sum_{\hat{a}\in
  A}\exp{b*Q_i^*(s,\hat{a}))}},
  \label{equation:loglikelihood}
\end{align}
where the optimal Q function is given by Equation \eqref{equation:approxQ}. After estimating the
parameter $\theta_i$, the value function and reward function can be computed with  Equation
\eqref{equation:approxrewardvalue}, \eqref{equation:approxV}, and \eqref{equation:approxR}. 

\subsection{Loss for Reward Sharing}
Since the demonstrator makes minimal changes to adapt the inherent reward function $r_b(s)$ into
task-specific one $r_i(s)$, we model the distribution as:
\[P(r_i(s)|r_b(s))\propto \exp{(\mathbb{D}(r_i(s),r_b(s))))}\]
where $\mathbb{D}(r_i(s),r_b(s)))$ measures the difference between $r_i(s)$ and $r_b(s)$. Thus the
loss function for reward sharing is given as:
\[L(r_i(s),r_b(s))=\log{Z}-\mathbb{D}(r_i(s),r_b(s)))\]
where $\log{Z}$ is the partition function and remains the same for all $r_i(s)$.

We test several functions as $\mathbb{D}(r_i(s),r_b(s)))$. The first choice is L2 loss, where
\[D(r_i(s),r_b(s))=||\{r_i(s)-r_b(s)\}||\]
where $\{r_i(s)-r_b(s)\}$ denotes the set of differences, evaluated on the full state space or only
the visited states.

The second choice is Huber loss with $\delta=1$, a differentiable approximation of the L1 loss
popular in sparse models:
\[L_{\delta }(a)={\begin{cases}{\frac {1}{2}}{a^{2}}&{\text{for }}|a|\leq \delta ,\\\delta
(|a|-{\frac {1}{2}}\delta ),&{\text{otherwise,}}\end{cases}},\]
and 
\[D(r_i(s),r_b(s))=\sum_{s}L_{1}(r_i(s)-r_b(s)).\] 

The third choice is standard deviation:
\[D(r_i(s),r_b(s))=stdev(\{r_i(s)-r_b(s)\})\]

The fourth choice is information entropy, after converting $\{r_i(s)-r_b(s)\}$ into a probabilistic
distribution with sofmax function:
\[D(r_i(s),r_b(s))=entropy(softmax(\{r_i(s)-r_b(s)\}))\]

With the loss function for IRL and reward sharing, the reward functions can be learned via gradient
method. The algorithm is shown in Algorithm \ref{alg:metairl}.
\begin{algorithm}[tb]
  \caption{Meta IRL}
  \label{alg:metairl}
\begin{algorithmic}[1]
  \STATE Data: {$\zeta,S,A,P,\gamma,b,\alpha$}
  \STATE Result: {reward value $R[S]$}
  \STATE create variable $\theta_b=\{W,b\}$ for a neural network
  \STATE create variable $\theta_i=\{W,b\},i=1,\cdots,\mathbb{T}$ for each task 
  \STATE initialize $\theta_b,\theta_i=\{W,b\},i=1,\cdots,\mathbb{T}$
  \WHILE{Not converging}
  \STATE update $\theta_b,\theta_i=\{W,b\},i=1,\cdots,\mathbb{T}$
 based on optimization \eqref{equation:metaoptimization}
  \ENDWHILE
  \STATE compute and return all $R_i[S]$
\end{algorithmic}
\end{algorithm}

\section{Experiments}
\label{irl::experiments}
\subsection{Path Planning}
We consider a path planning problem on an uneven terrain, where an agent can observe the whole
terrain to find the optimal paths from random starting points to arbitrary goal points, but a mobile
robot can only observe the agent's demonstrations to learn how to plan paths. Given a starting point
and a goal point, an optimal path depends solely on the costs to move across the terrain. To learn
the costs, we formulate a Markov Decision Process for each goal point, where a state denotes a small
region of the terrain and an action denotes a possible movement. The reward of a state equals to the
negative of the cost to move across the corresponding region, while the goal state has an additional
reward to attract movements. 

In this work, we create a discretized terrain with several hills, where each hill is defined as a
peak of cost distribution and the costs around each hill decay exponentially, and the true cost of a
region is the summation of the costs from all hills. Ten worlds are randomly generated, and in each
world, ten tasks are generated, each with a different goal state. For each task, the agent
demonstrates ten trajectories, where the length of a trajectory depends on how many steps to reach
the goal state.

\begin{figure*}
  \centering
  \subfloat[1 task]{\includegraphics[width=0.23\textwidth]{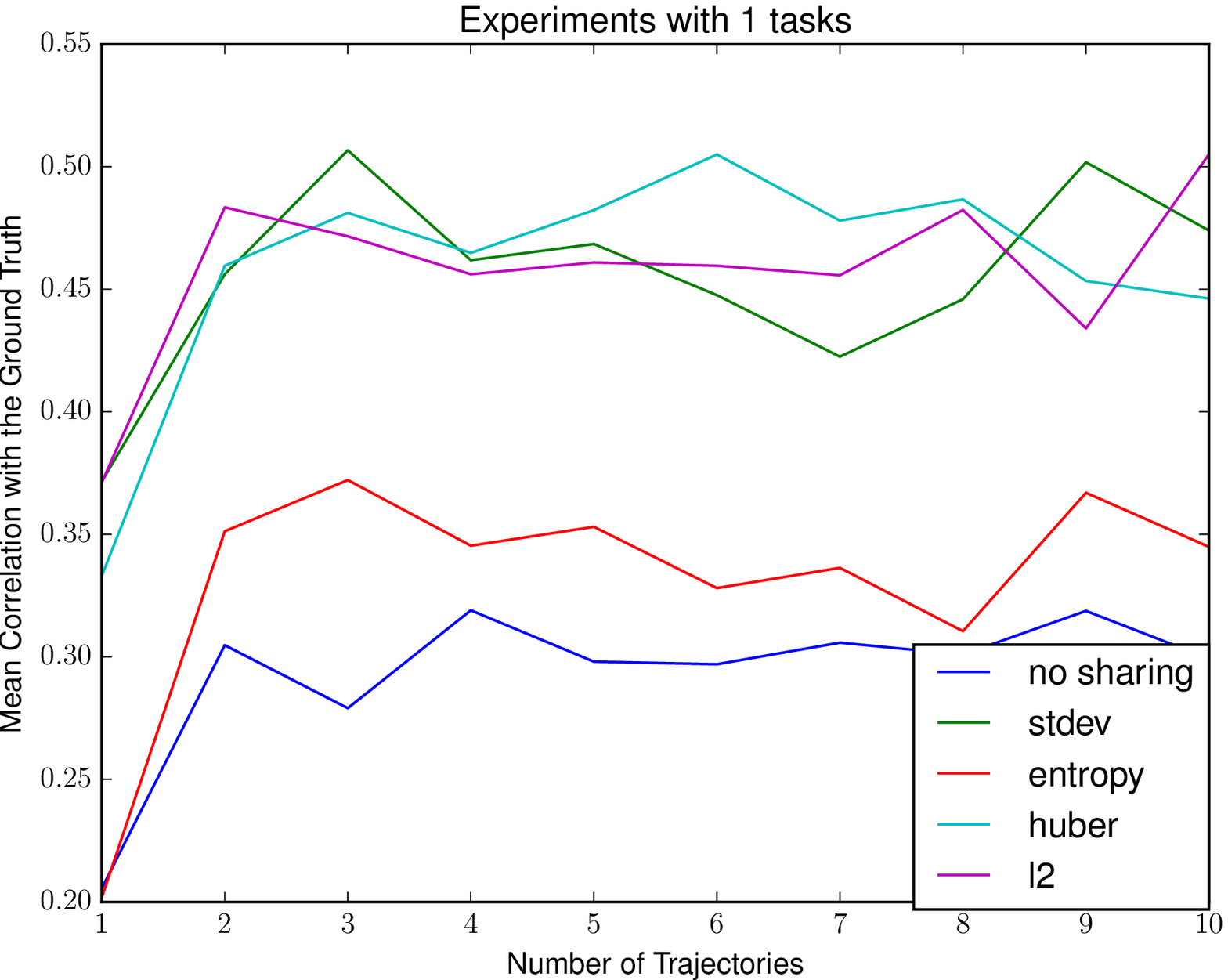}\label{fig:result1}}
  \subfloat[2 tasks]{\includegraphics[width=0.23\textwidth]{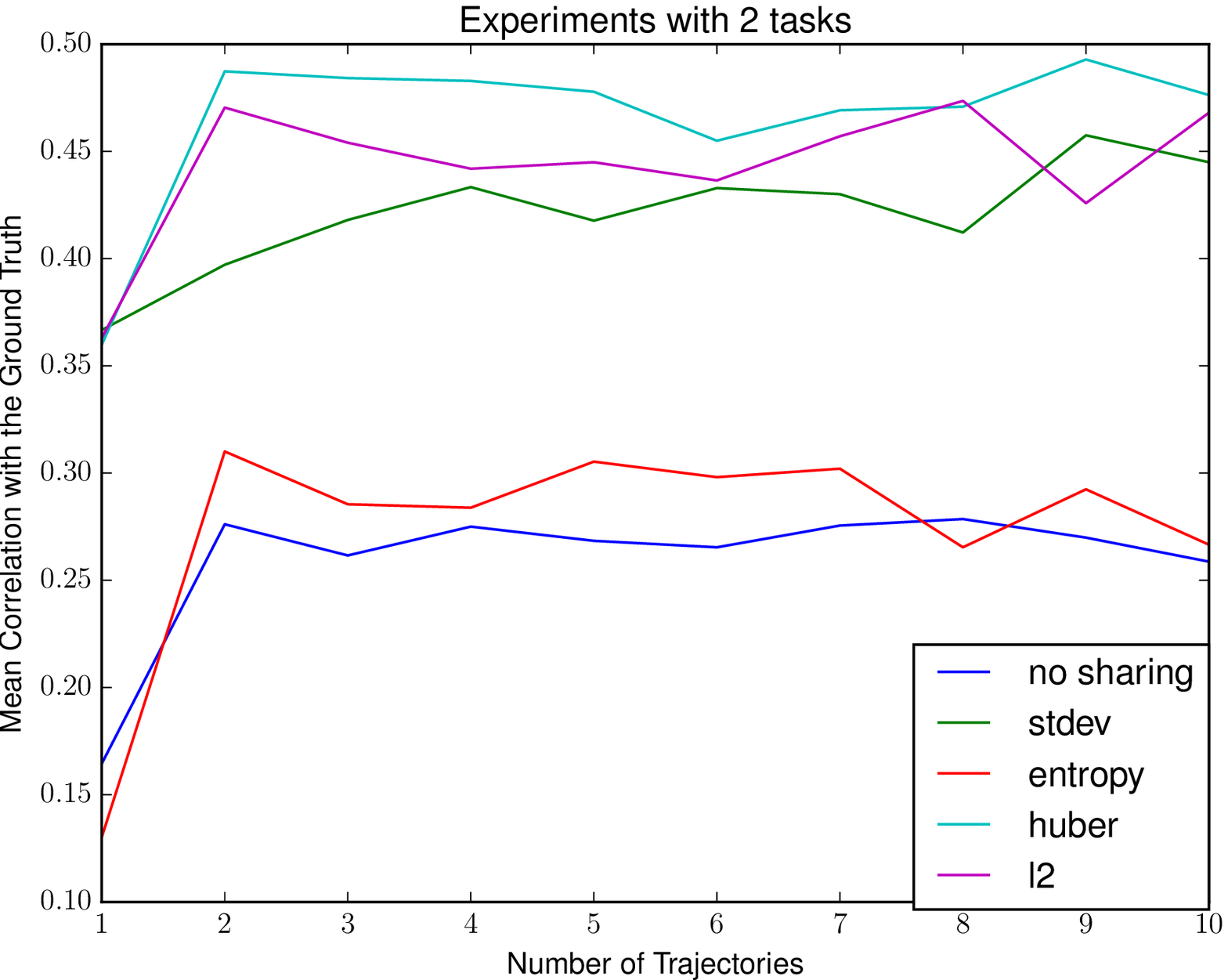}\label{fig:result2}}
  \subfloat[3 tasks]{\includegraphics[width=0.23\textwidth]{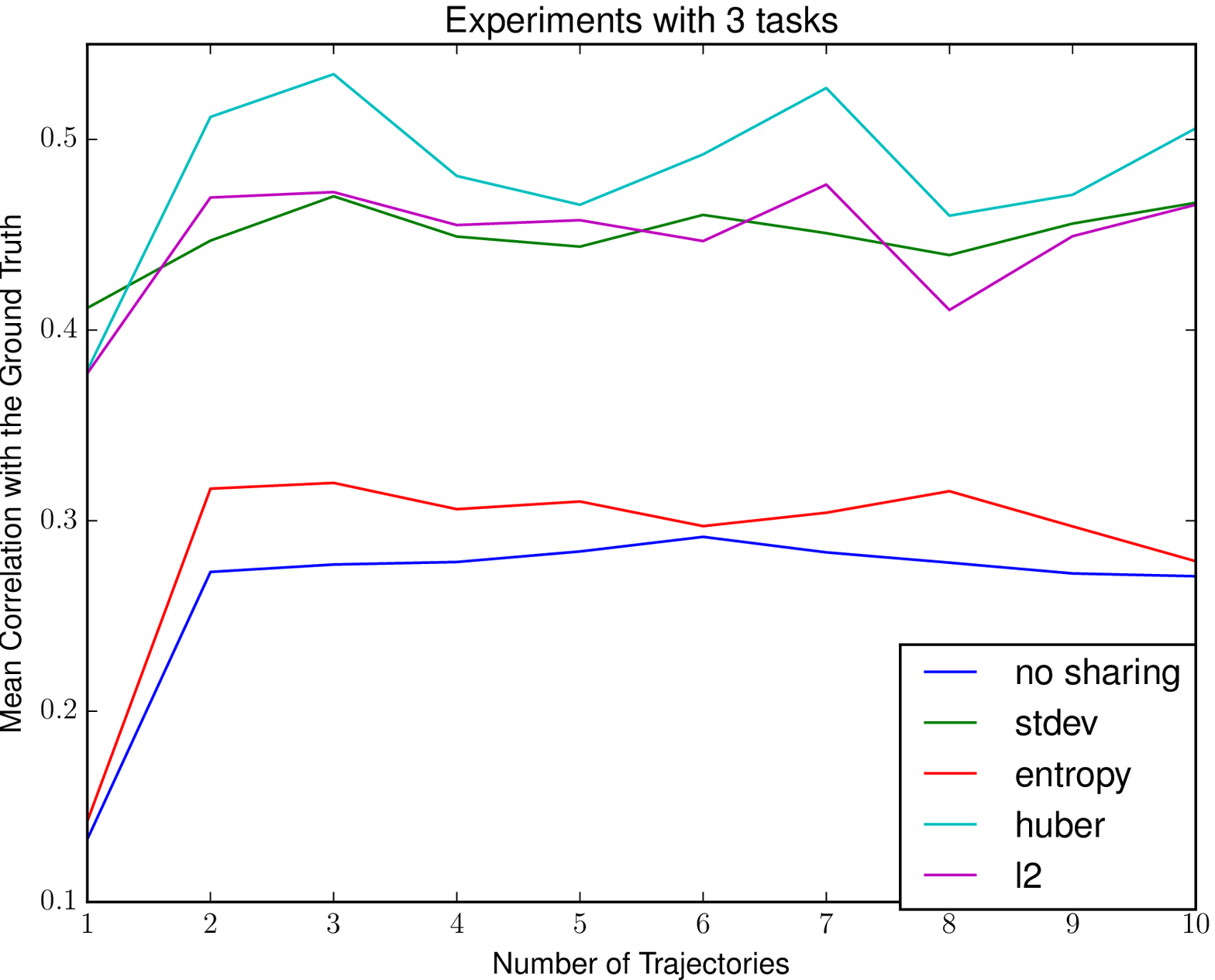}\label{fig:result3}}
  \subfloat[4 tasks]{\includegraphics[width=0.23\textwidth]{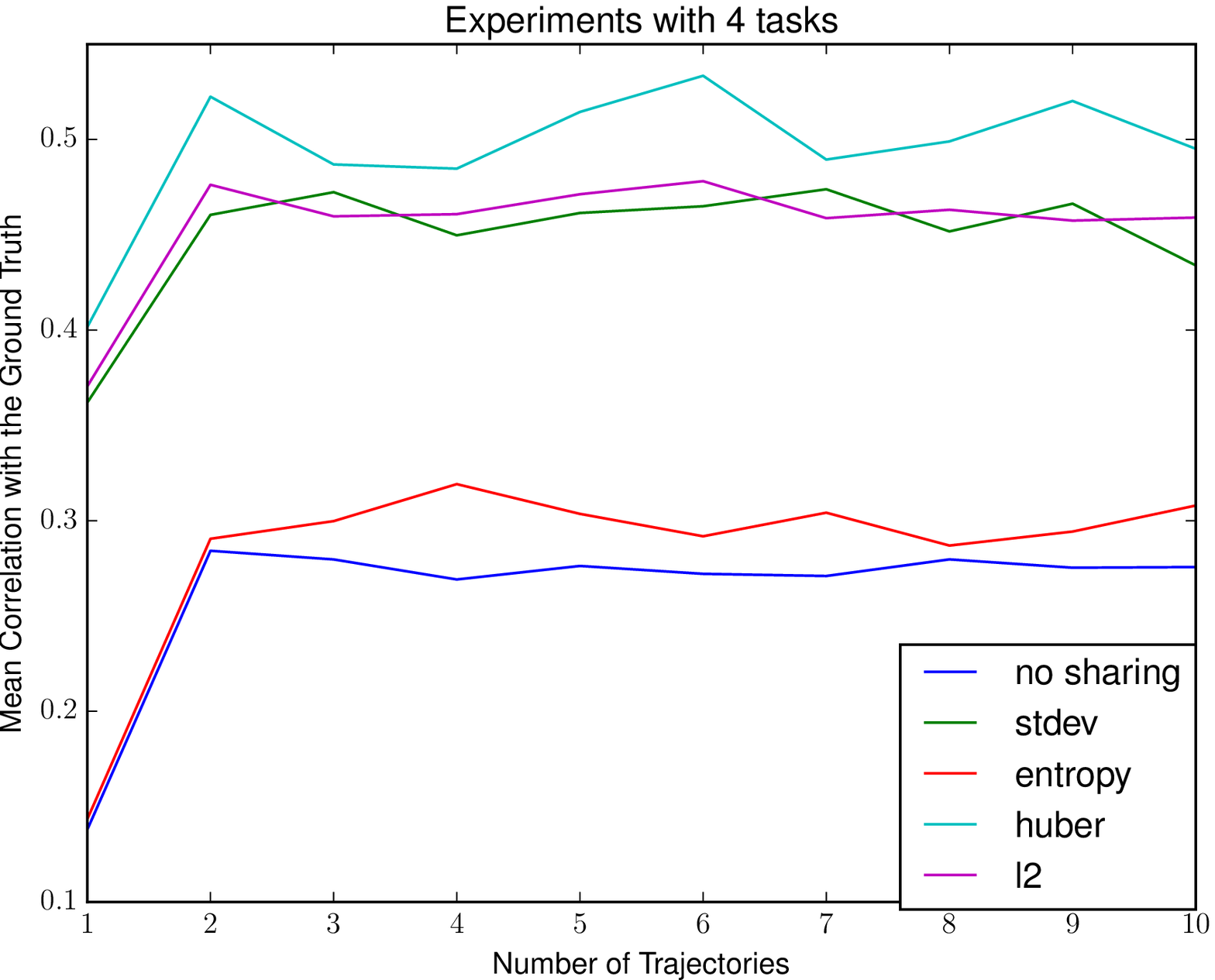}\label{fig:result4}}

  \subfloat[5 tasks]{\includegraphics[width=0.23\textwidth]{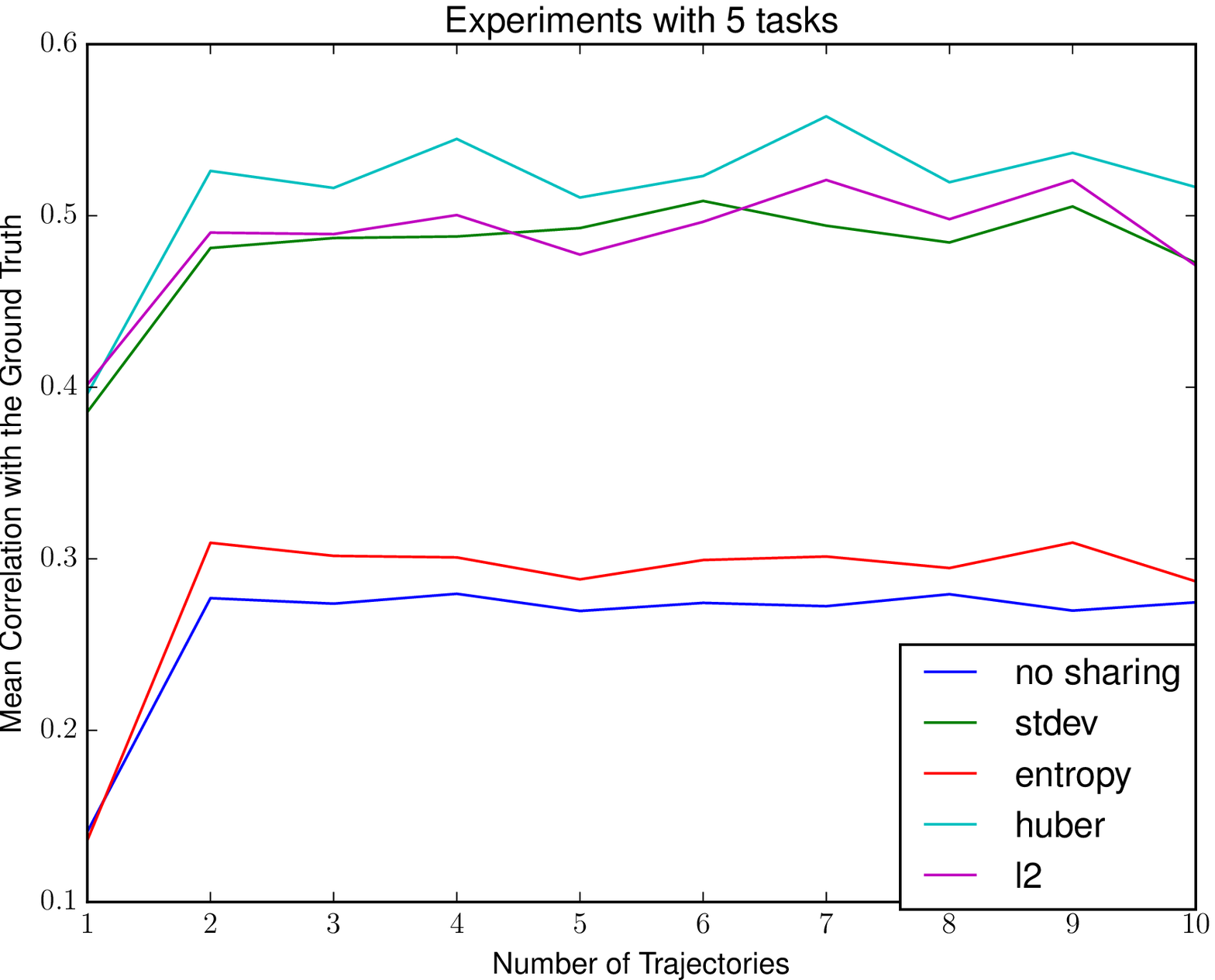}\label{fig:result5}}
  \subfloat[6 tasks]{\includegraphics[width=0.23\textwidth]{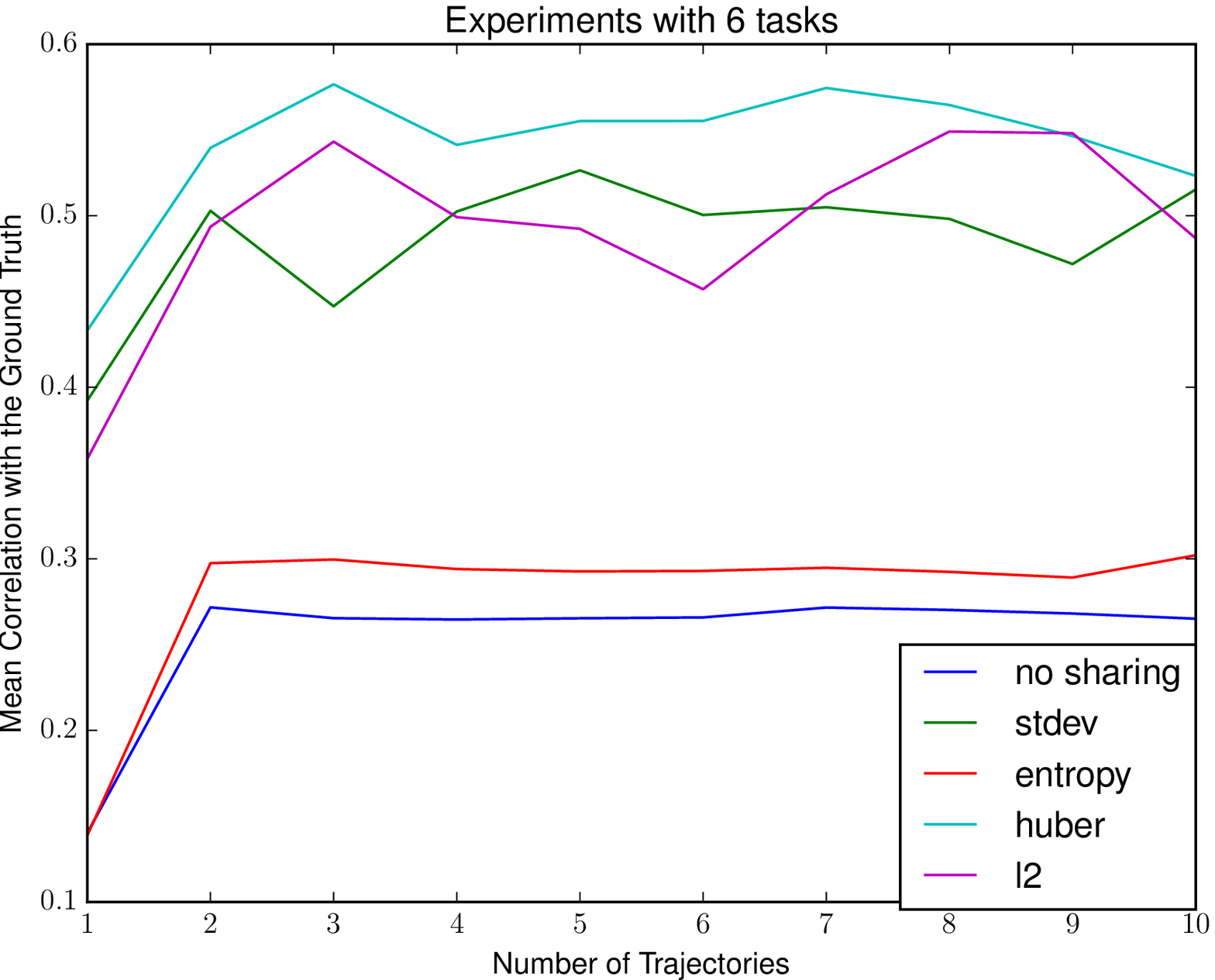}\label{fig:result6}}
  \subfloat[7 tasks]{\includegraphics[width=0.23\textwidth]{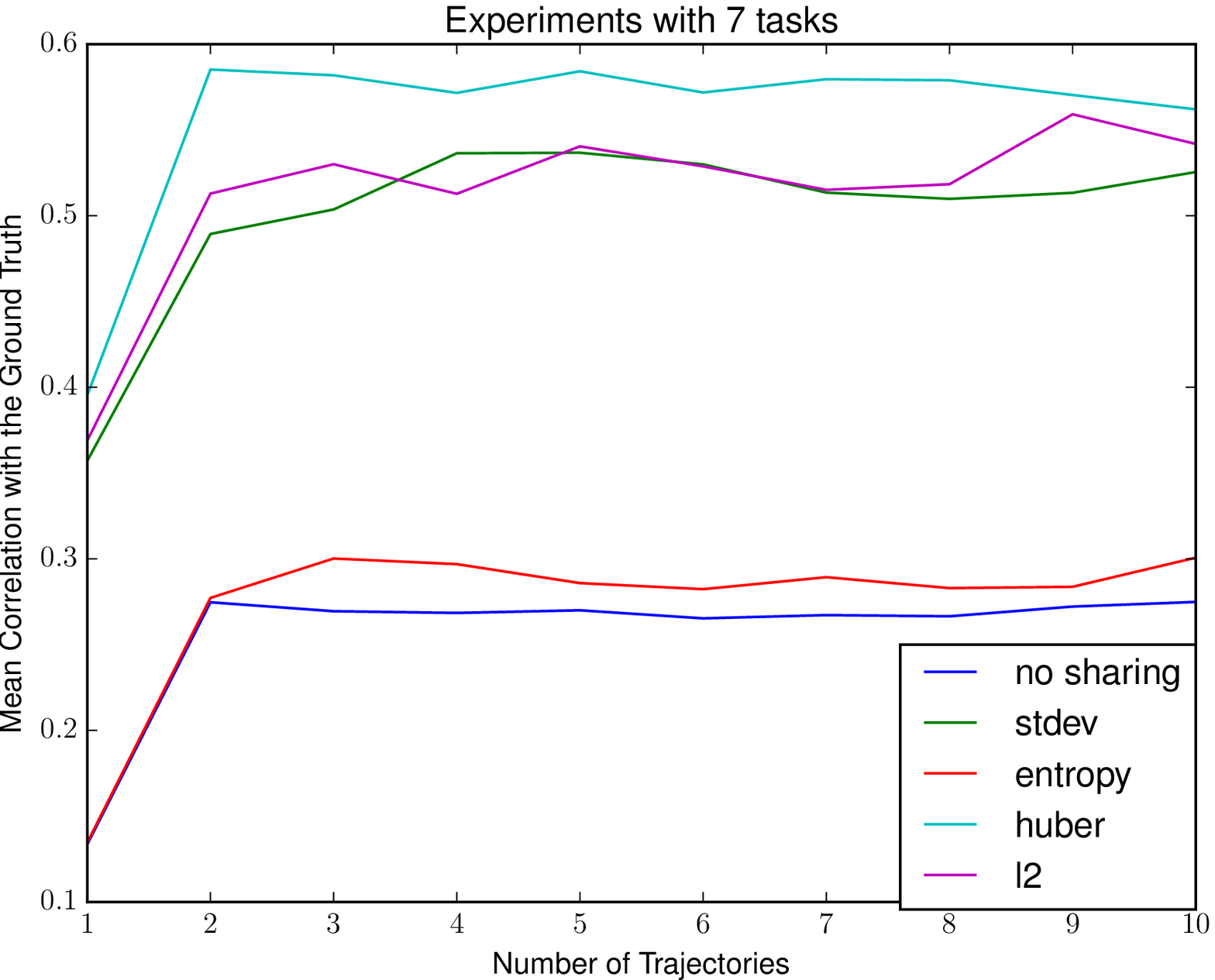}\label{fig:result7}}
  \subfloat[8 tasks]{\includegraphics[width=0.23\textwidth]{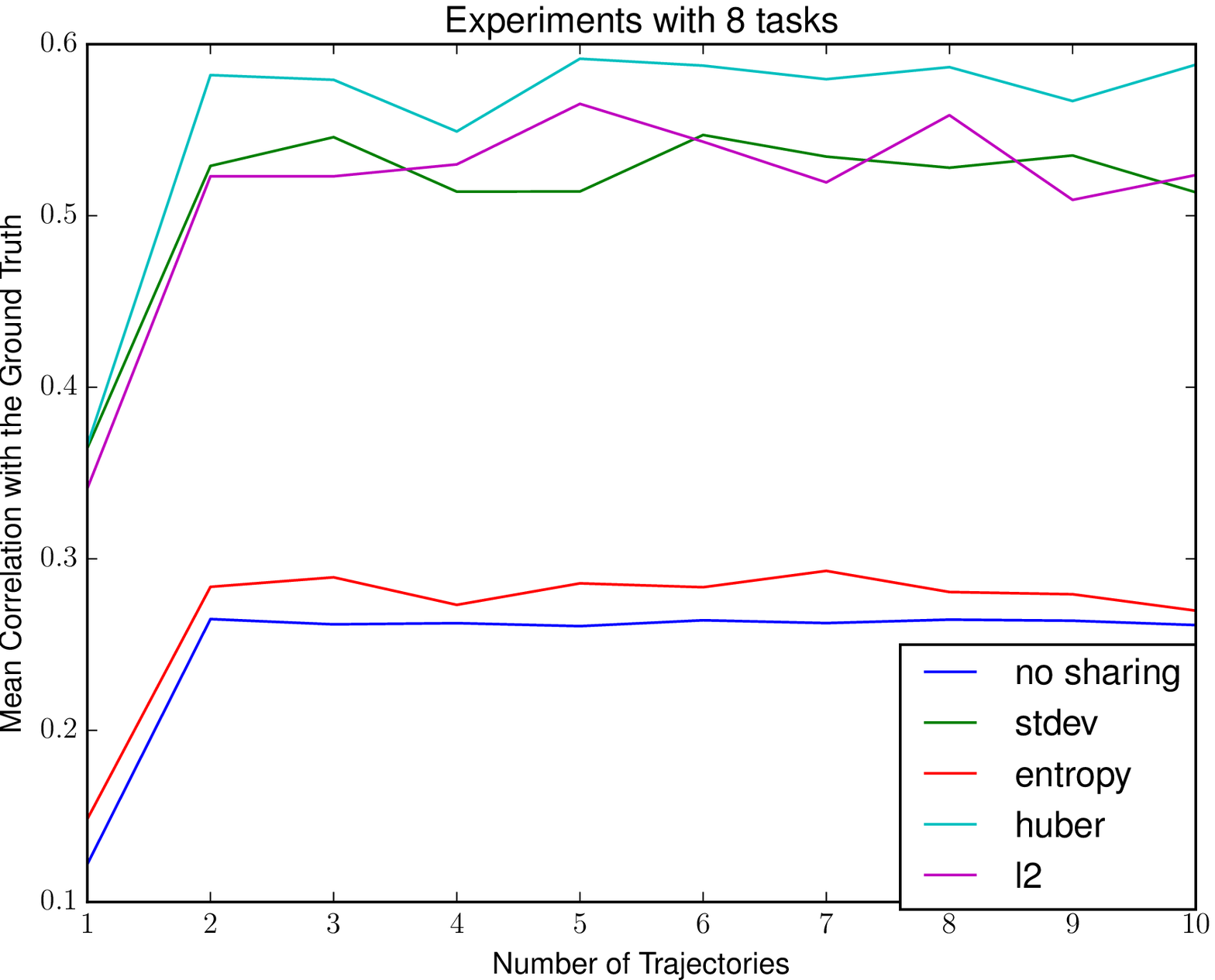}\label{fig:result8}}

  \subfloat[9 tasks]{\includegraphics[width=0.23\textwidth]{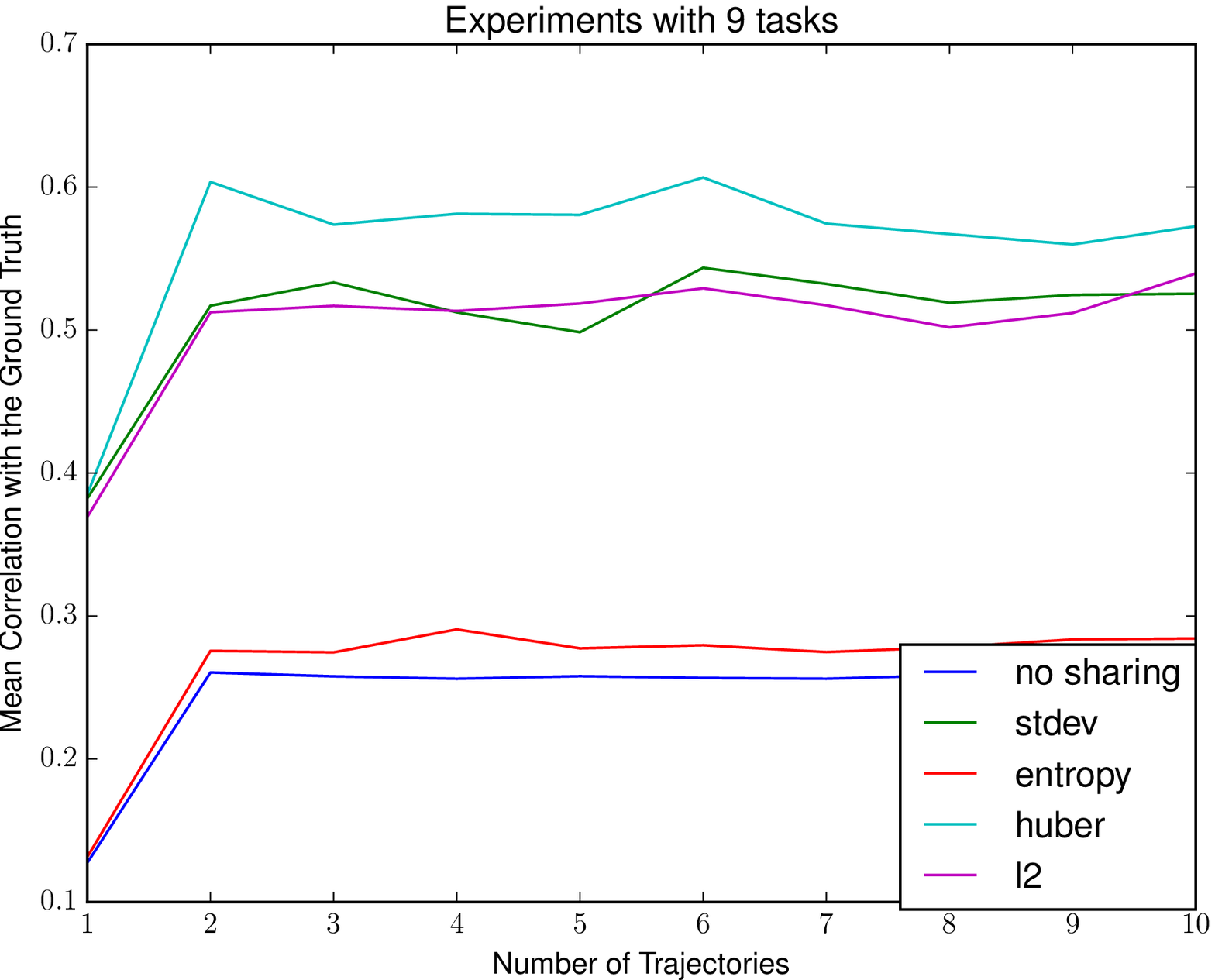}\label{fig:result9}}
  \subfloat[10 tasks]{\includegraphics[width=0.23\textwidth]{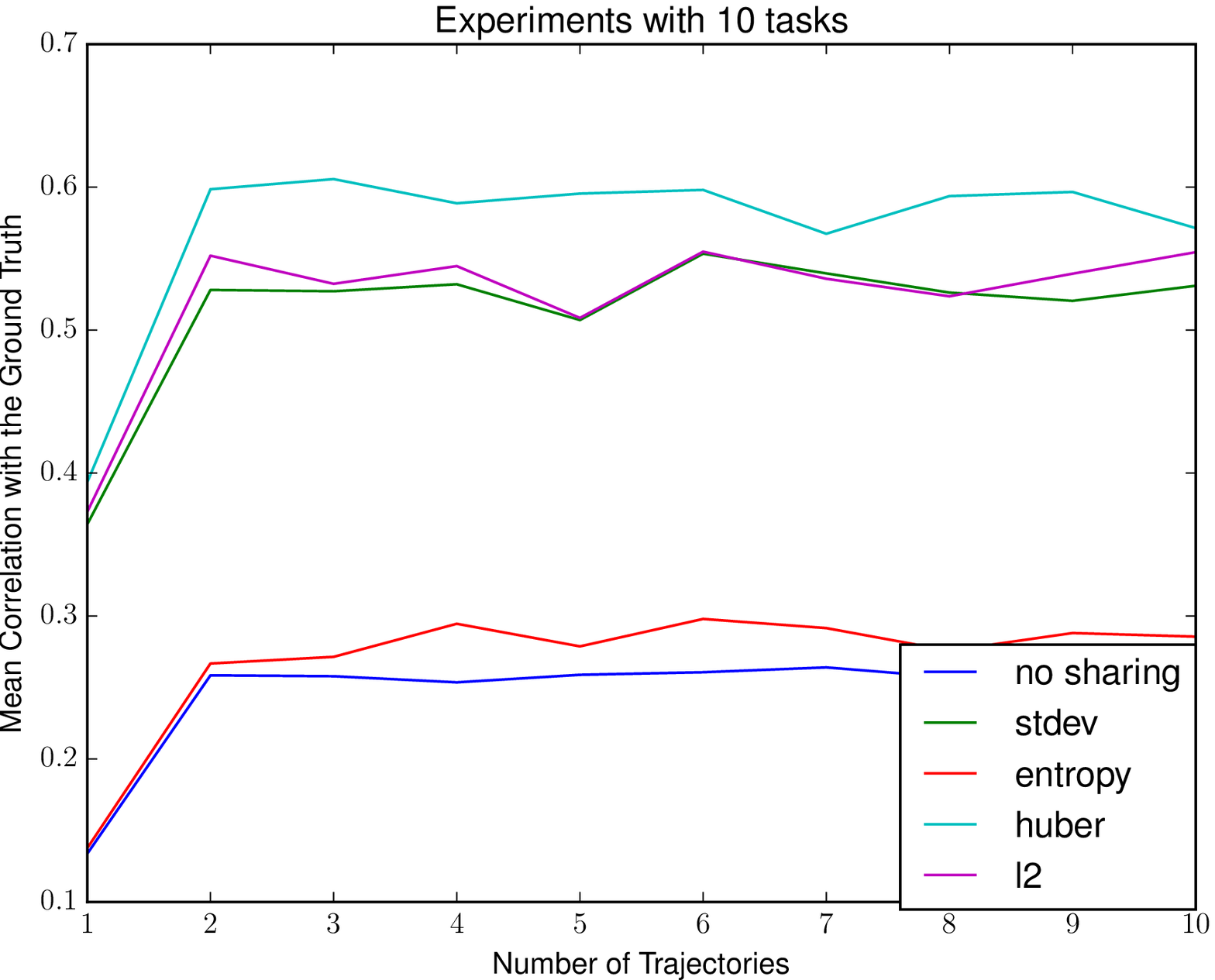}\label{fig:result10}}
  \subfloat[11 tasks]{\includegraphics[width=0.23\textwidth]{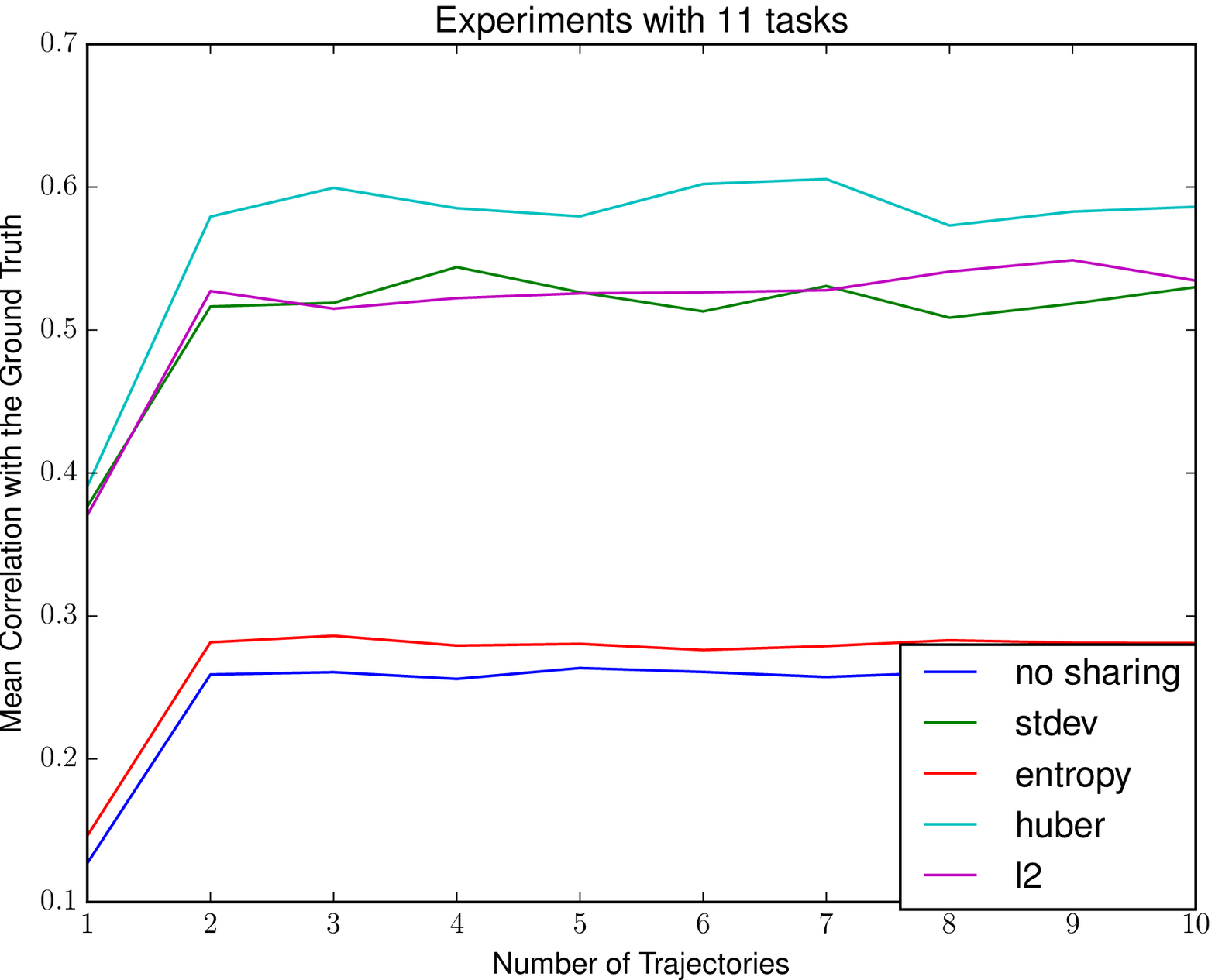}\label{fig:result11}}
  \subfloat[12 tasks]{\includegraphics[width=0.23\textwidth]{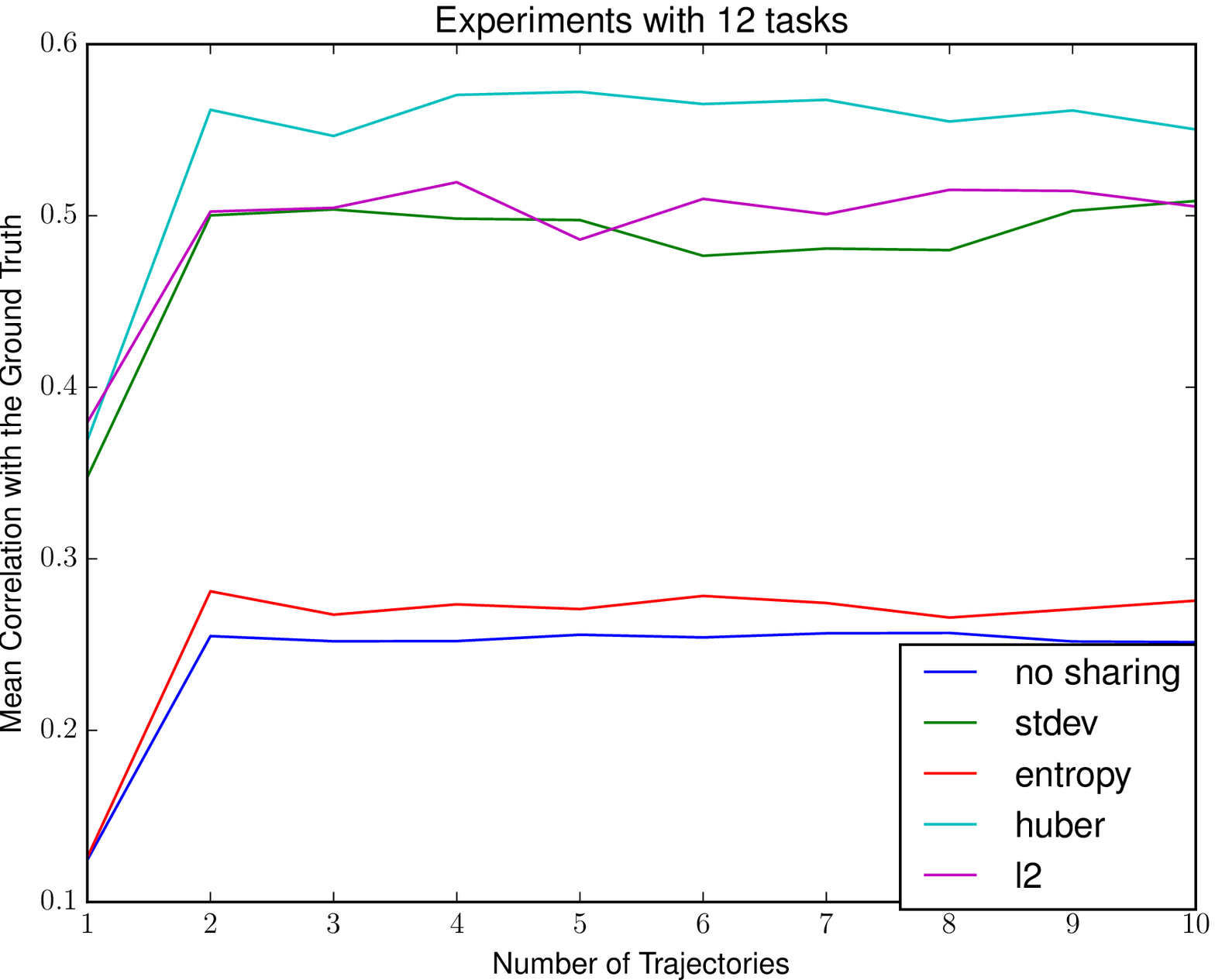}\label{fig:result12}}

  \subfloat[13 tasks]{\includegraphics[width=0.23\textwidth]{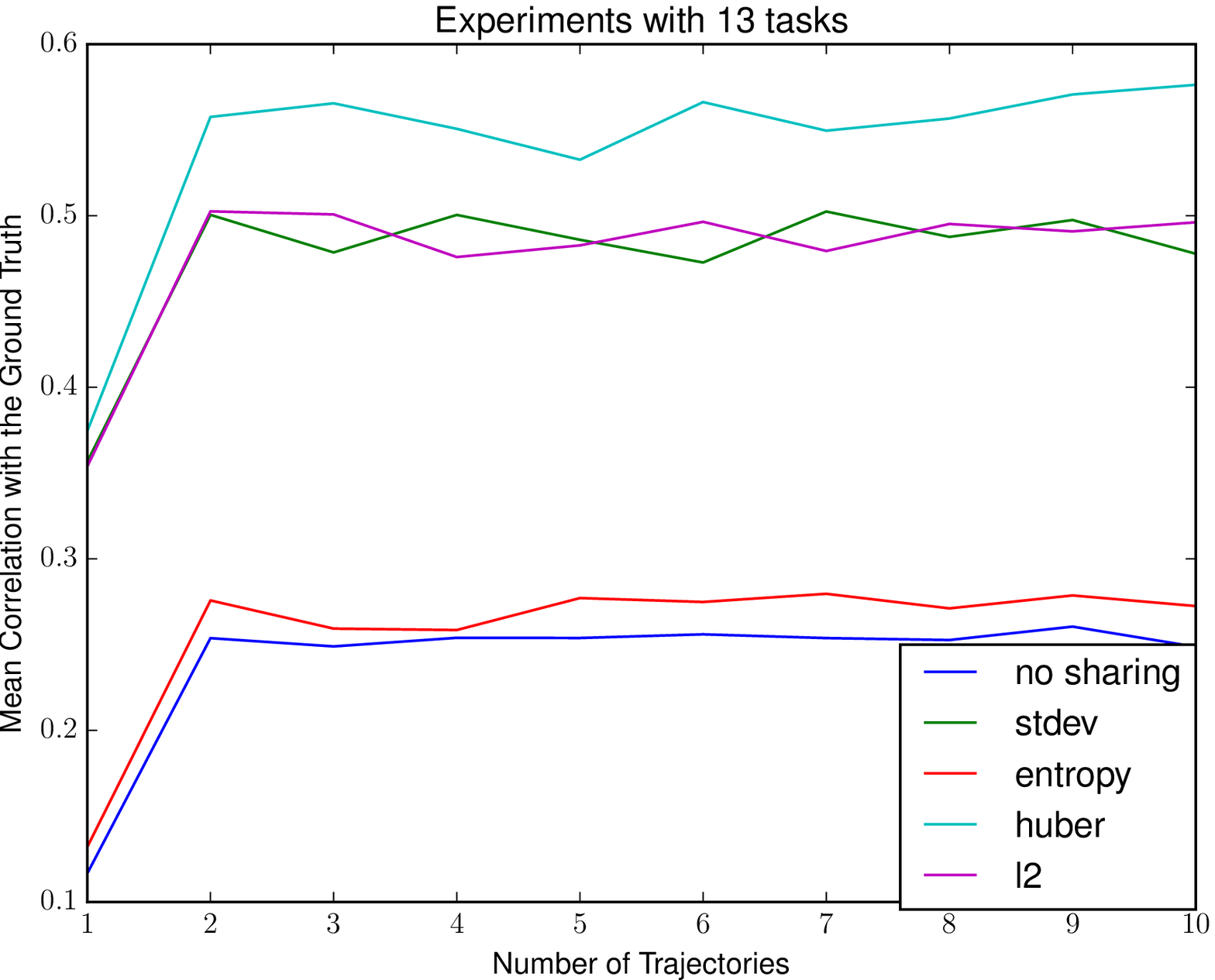}\label{fig:result13}}
  \subfloat[14 tasks]{\includegraphics[width=0.23\textwidth]{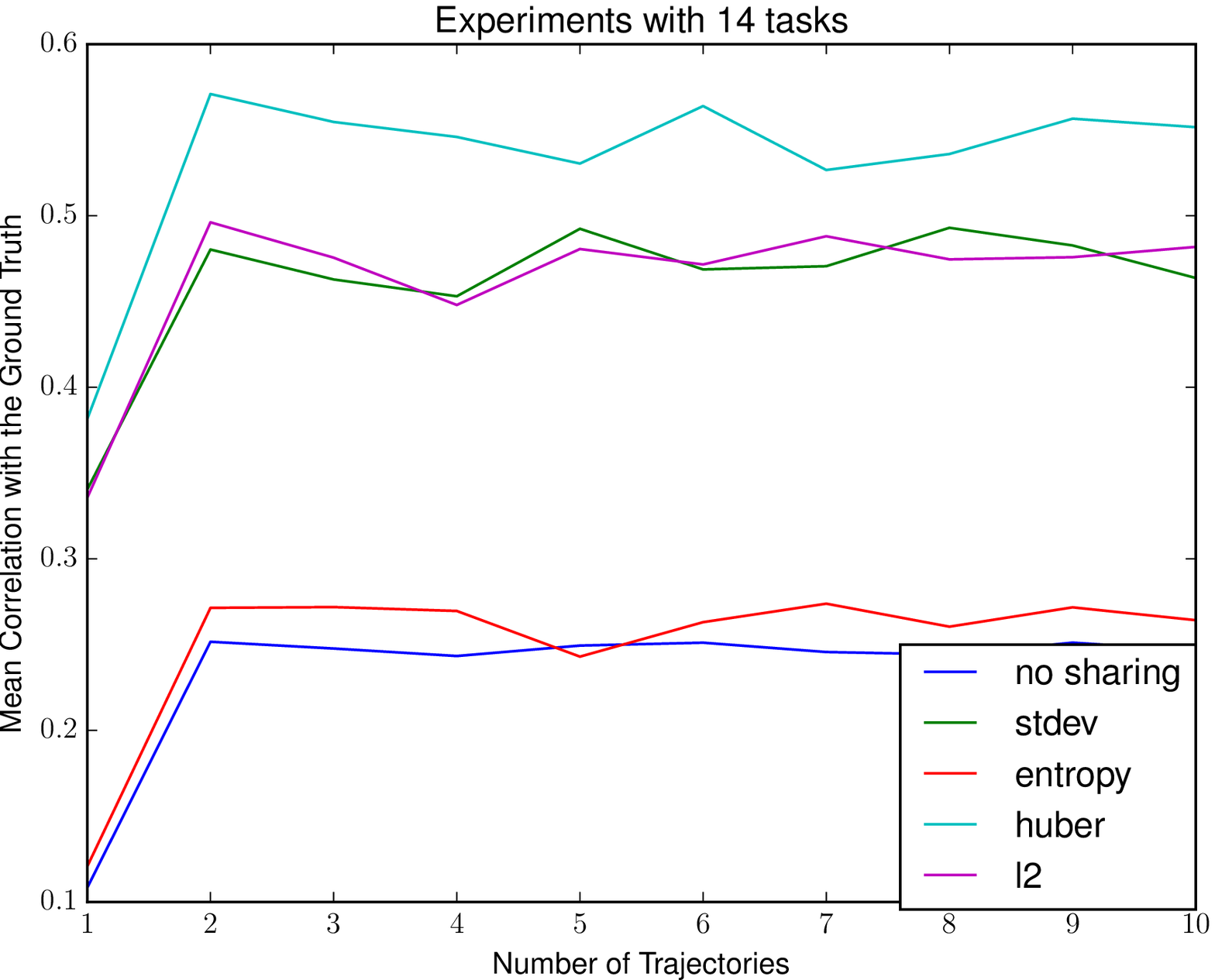}\label{fig:result14}}
  \subfloat[15 tasks]{\includegraphics[width=0.23\textwidth]{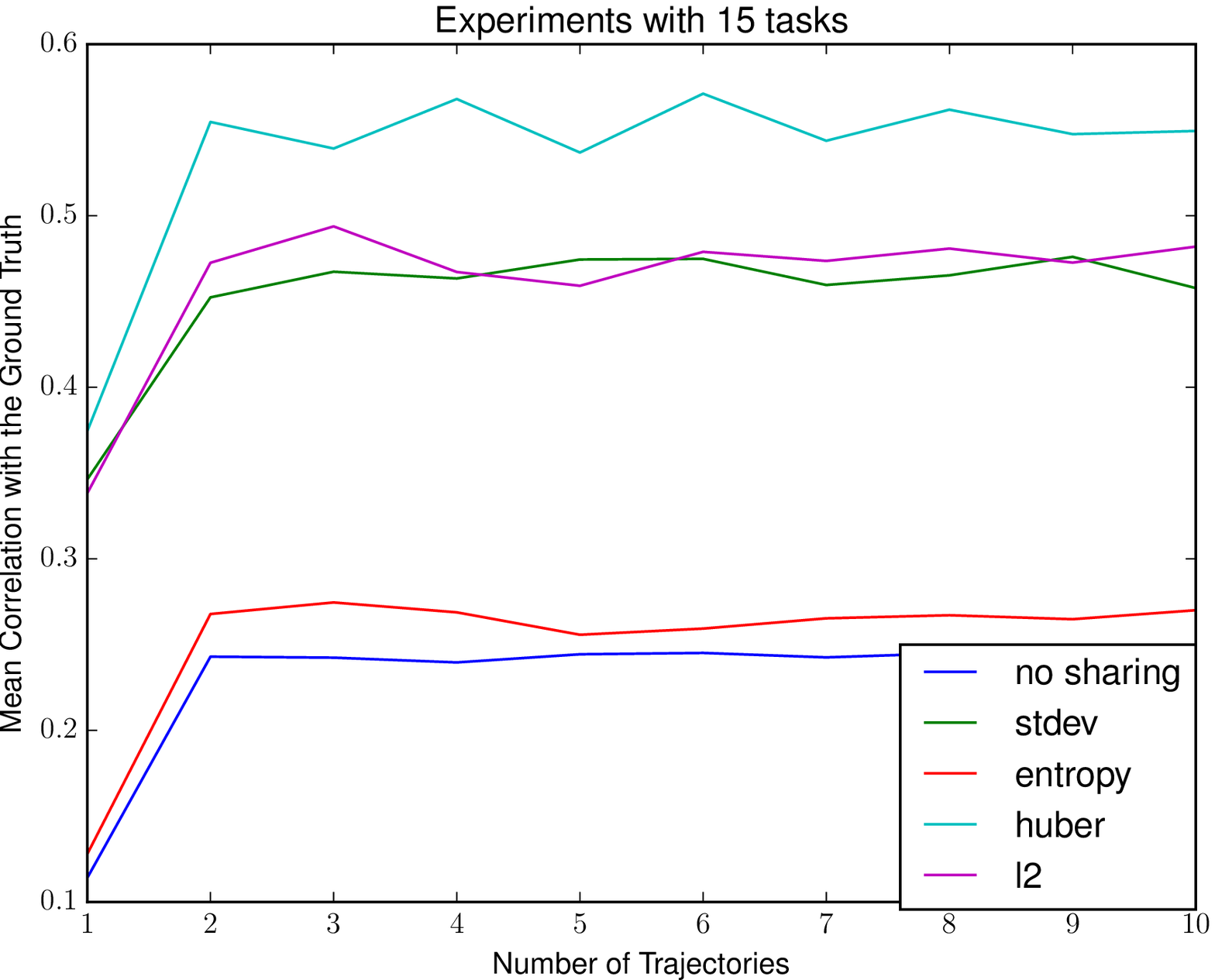}\label{fig:result15}}
  \subfloat[16 tasks]{\includegraphics[width=0.23\textwidth]{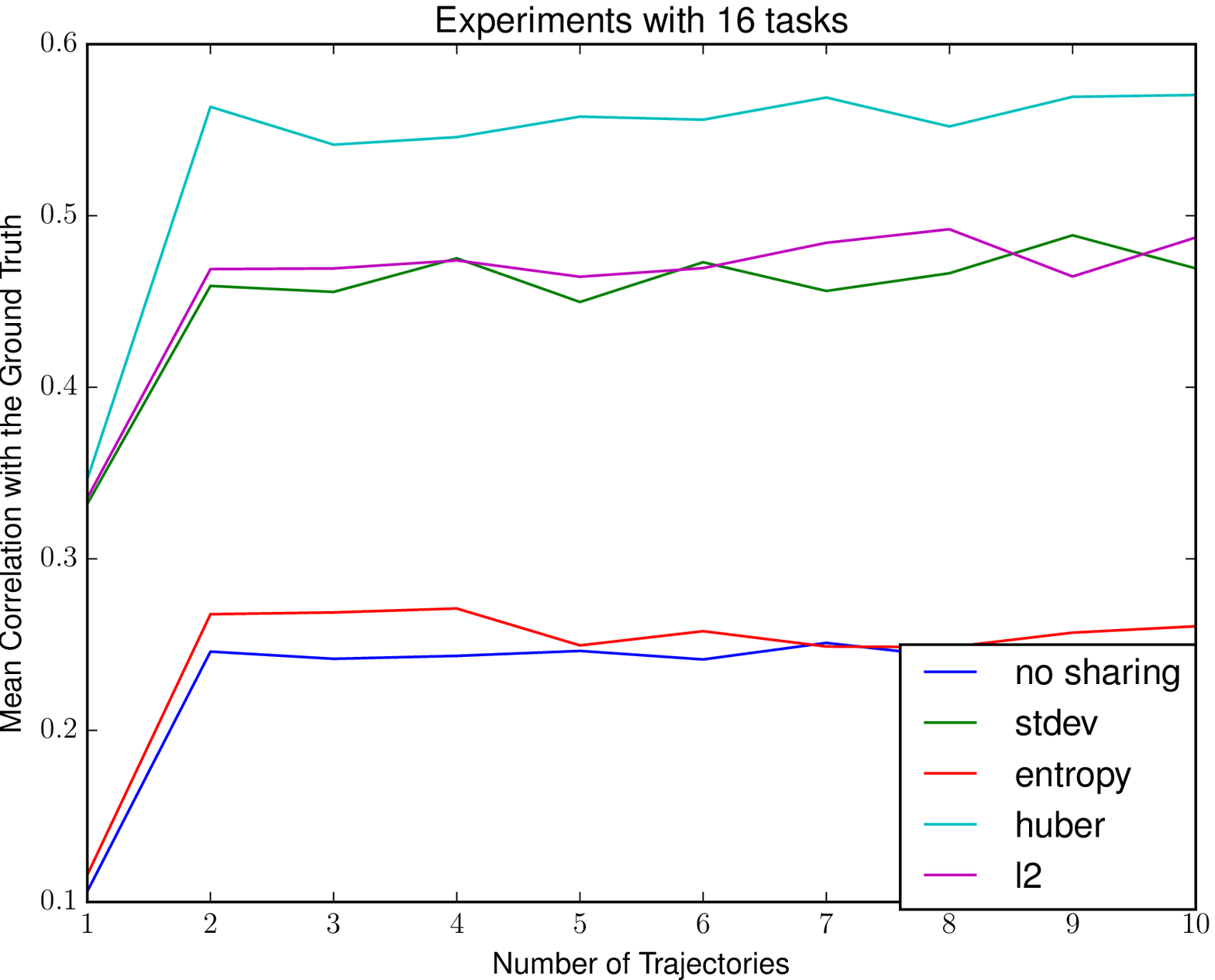}\label{fig:result16}}
  \caption{The result with five reward sharing loss functions on 16 tasks with
  at most 10 demonstrations for each task in 10 environment.}
  \label{fig:metaresult}

\end{figure*}

We evaluate the proposed method with different reward sharing loss functions under different number
of tasks and different number of trajectories. The evaluated loss functions include no reward
sharing, reward sharing with standard deviation, information entropy, L2 loss, and huber loss. The
number of tasks ranges from 1 to 16, and for each task, the number of trajectories ranges from 1 to
10. The learning rate is 0.01, with Adam optimizer. The accuracy of a reward is computed as the
correlation coefficient between the learned reward function and the ground truth one. The results
are shown in Figure \ref{fig:metaresult}.

The result shows that the meta learning step can significantly improve the accuracy of reward
learning, among which the huber loss function leads to the best performance in average. L2 loss and
standard deviation have similar performance, not surprisingly. However, the information entropy has
a really bad performance.

\comment{
\subsection{Motion Analysis}
During rehabilitation, a patient with spinal cord injuries sits on a box, with a flat plate force
sensor mounted on box to capture the center-of-pressure (COP) of the patient during movement. Each
experiment is composed of two sessions, one without transcutaneous stimulation and one with
stimulation. The electrodes configuration and stimulation signal pattern are manually selected by
the clinician \cite{irl::spi}.

In each session, the physician gives eight (or four) directions for the patient to follow, including
left, forward left, forward, forward right, right, right backward, backward, backward left, and the
patient moves continuously to follow the instruction. The physician observes the patient's behaviors
and decides the moment to change the instruction.

Six experiments are done, each with two sessions. The COP trajectories in Figure \ref{fig:patient1}
denote the case with four directional instructions; Figure \ref{fig:patient2}, \ref{fig:patient3},
\ref{fig:patient4}, \ref{fig:patient5}, and \ref{fig:patient6} denote the sessions with eight
directional instructions.
\begin{figure}
  \centering
  \includegraphics[width=0.4\textwidth]{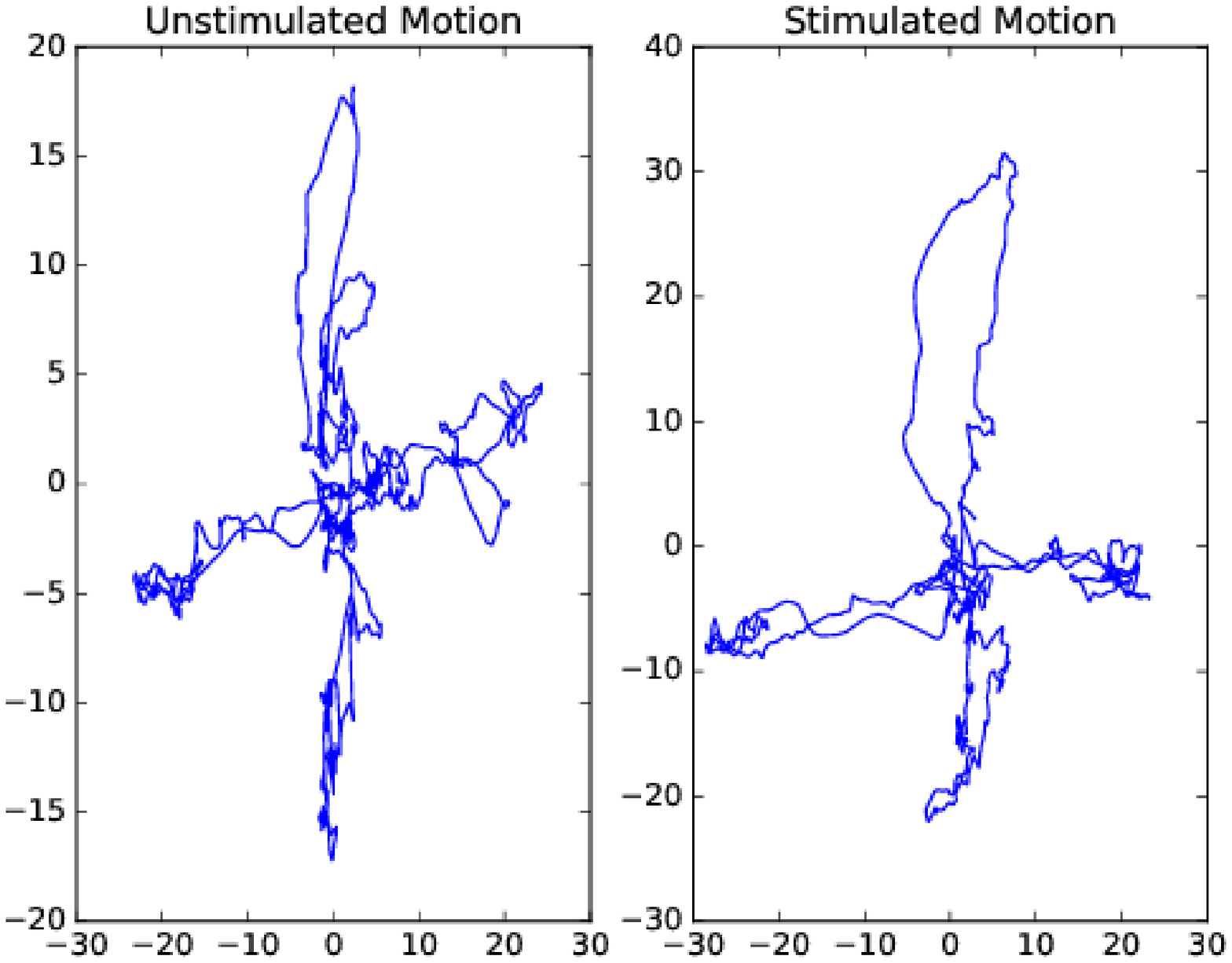}
  \caption{Patient 1 under four directional instructions: "unstimulated motion" means that the
  patient moves without transcutaneous stimulations, while "stimulated motion" represents the motion
  under stimulations.}
  \label{fig:patient1}
  \vspace{-0.5cm}
\end{figure}

\begin{figure}
  \centering
  \includegraphics[width=0.4\textwidth]{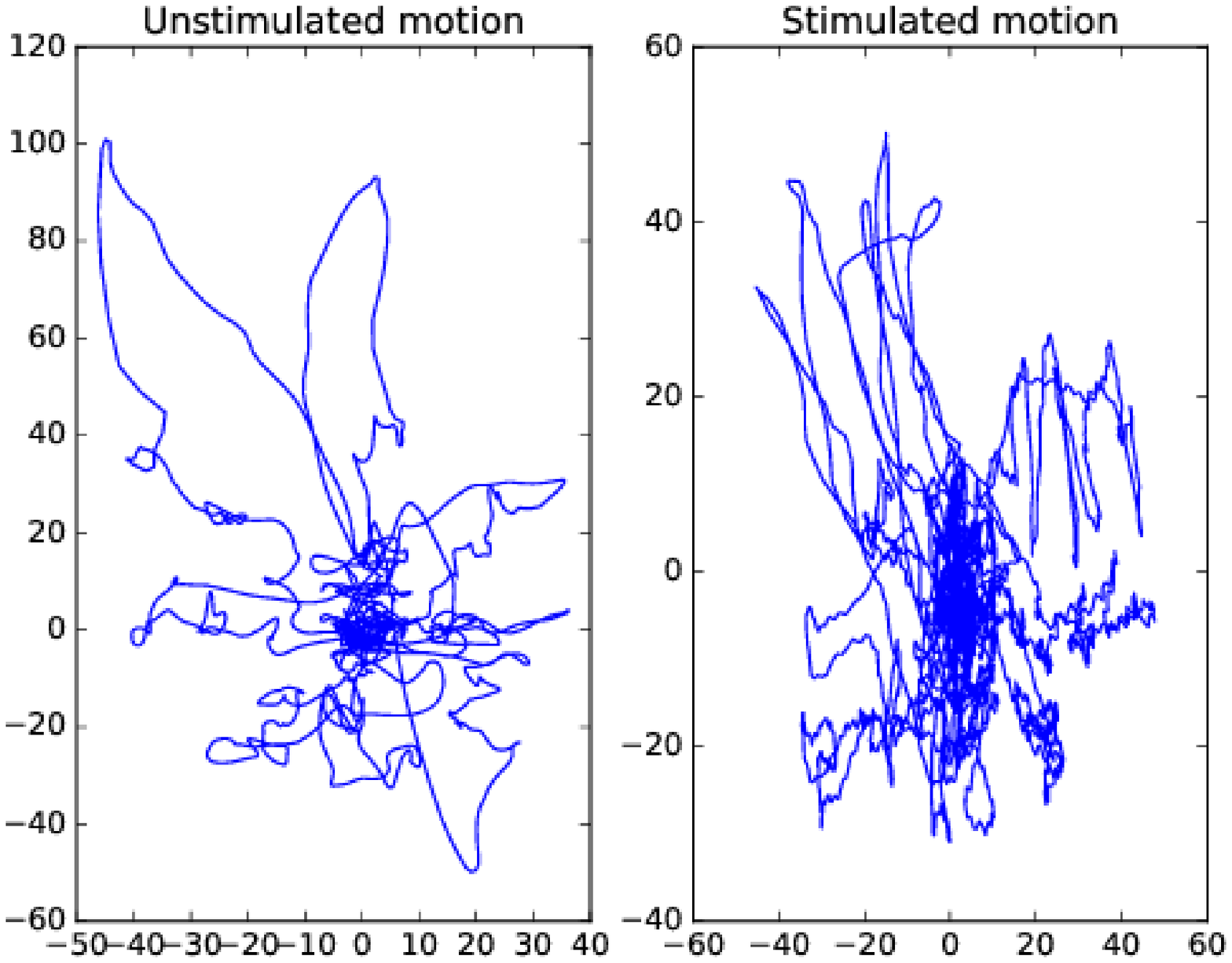}
  \caption{Patient 2 under eight directional instructions: "unstimulated motion" means that the
  patient moves without transcutaneous stimulations, while "stimulated motion" represents the motion
  under stimulations.}
  \label{fig:patient2}
  \vspace{-0.5cm}

\end{figure}
\begin{figure}
  \centering
  \includegraphics[width=0.4\textwidth]{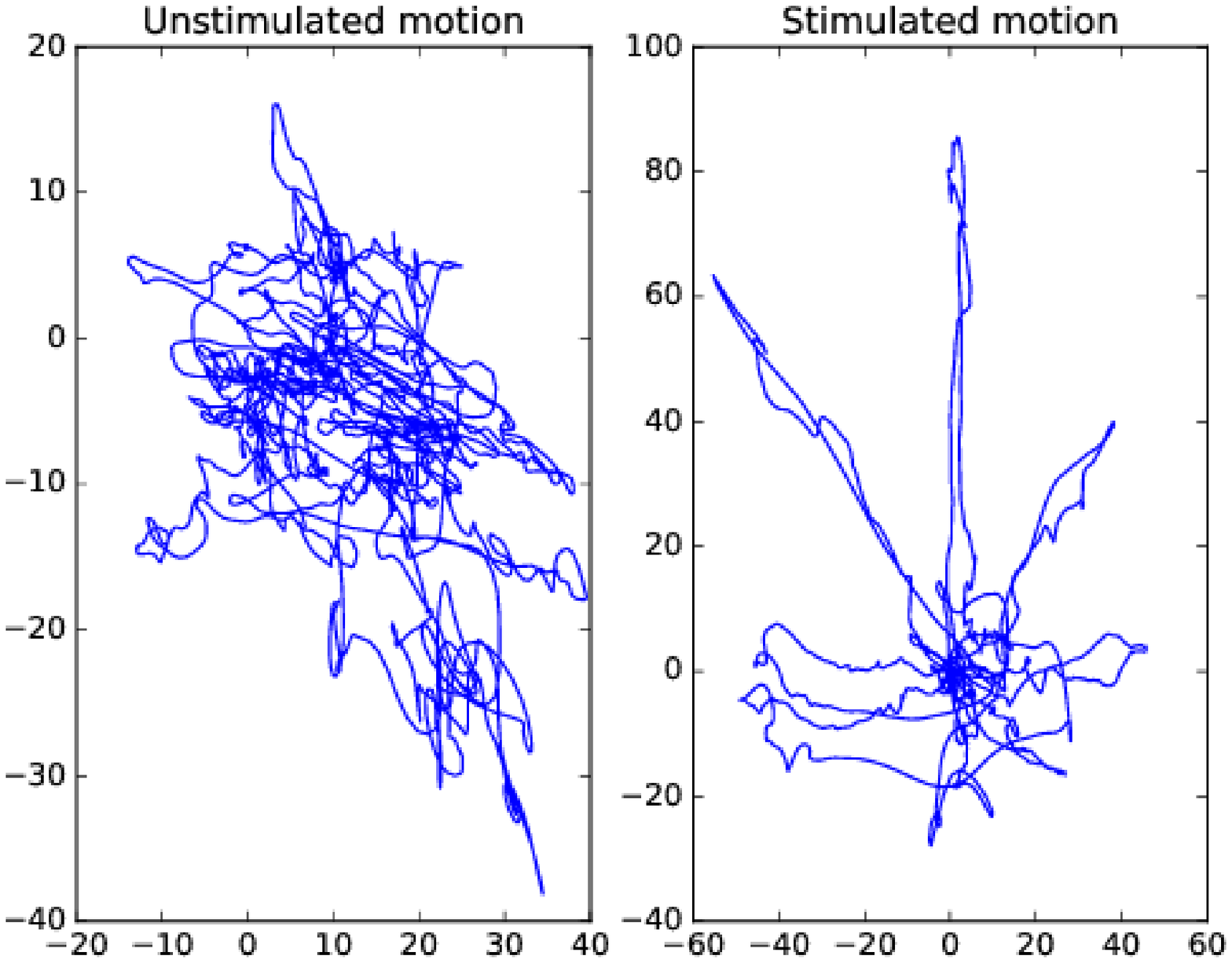}
  \caption{Patient 3 under eight directional instructions: "unstimulated motion" means that the
  patient moves without transcutaneous stimulations, while "stimulated motion" represents the motion
  under stimulations.}
  \label{fig:patient3}
  \vspace{-0.5cm}

\end{figure}
\begin{figure}
  \centering
  \includegraphics[width=0.4\textwidth]{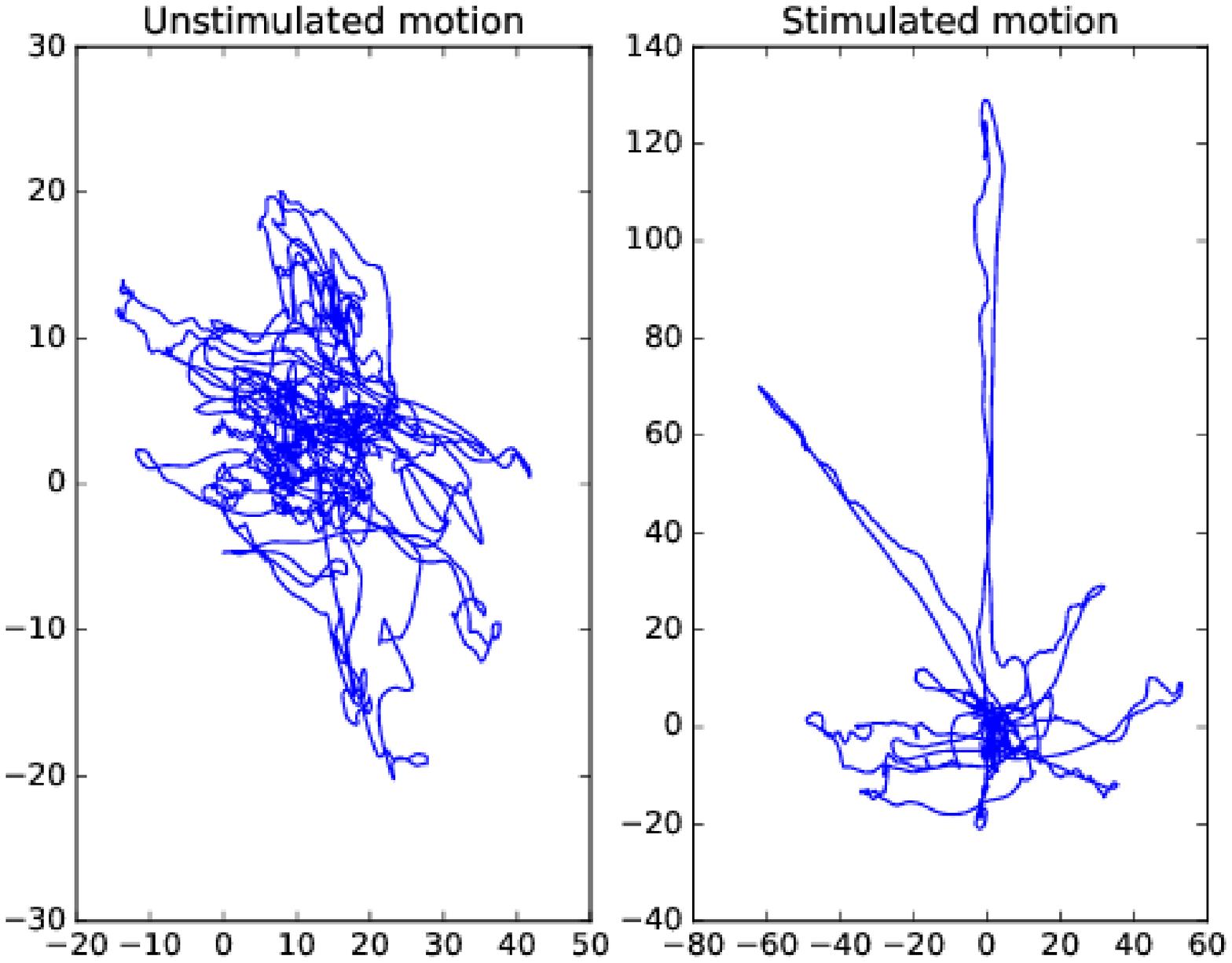}
  \caption{Patient 4 under eight directional instructions: "unstimulated motion" means that the
  patient moves without transcutaneous stimulations, while "stimulated motion" represents the motion
  under stimulations.}
  \label{fig:patient4}
  \vspace{-0.5cm}

\end{figure}
\begin{figure}
  \centering
  \includegraphics[width=0.4\textwidth]{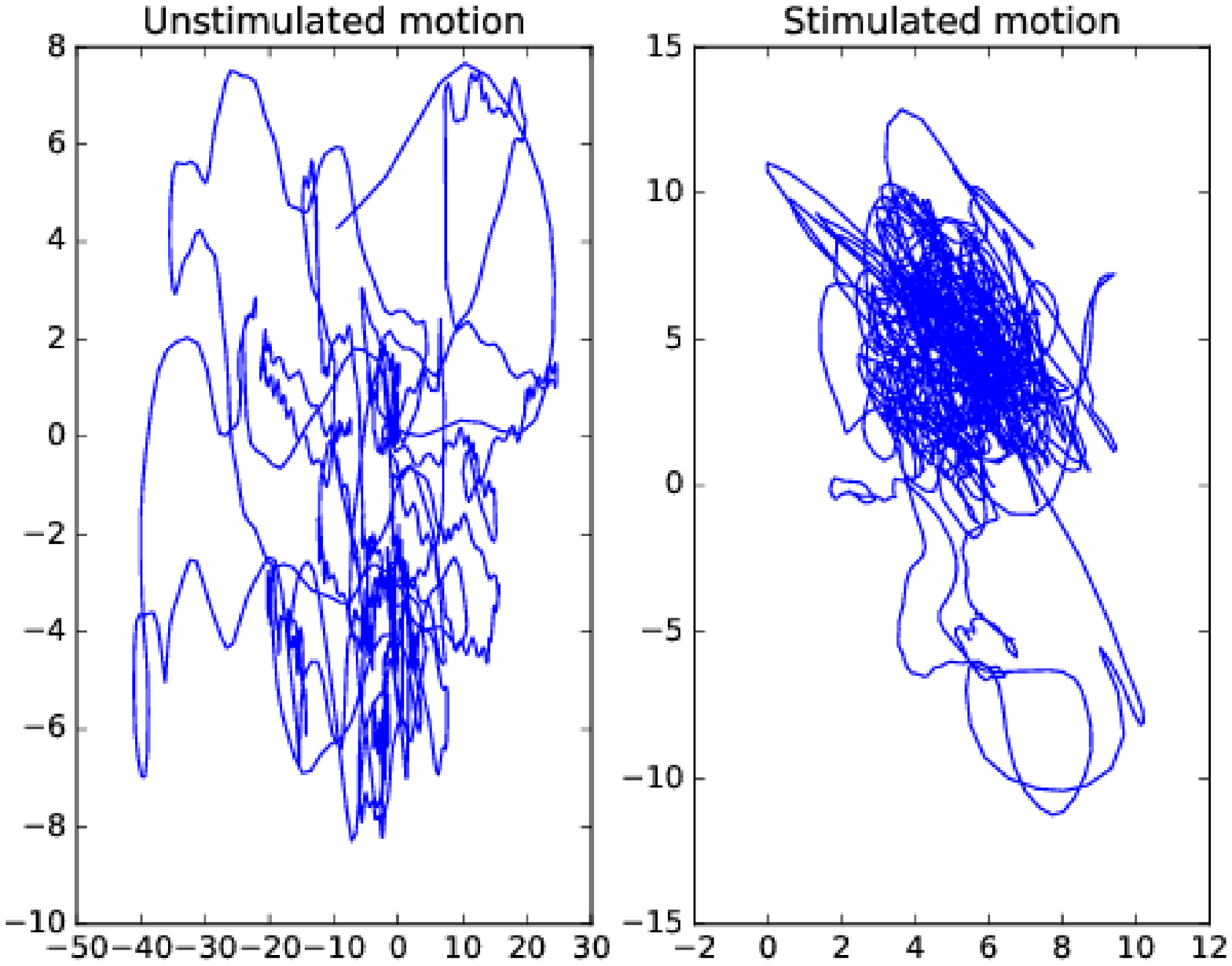}
  \caption{Patient 5 under eight directional instructions: "unstimulated motion" means that the
  patient moves without transcutaneous stimulations, while "stimulated motion" represents the motion
  under stimulations.}
  \label{fig:patient5}
  \vspace{-0.5cm}

\end{figure}
\begin{figure}
  \centering
  \includegraphics[width=0.4\textwidth]{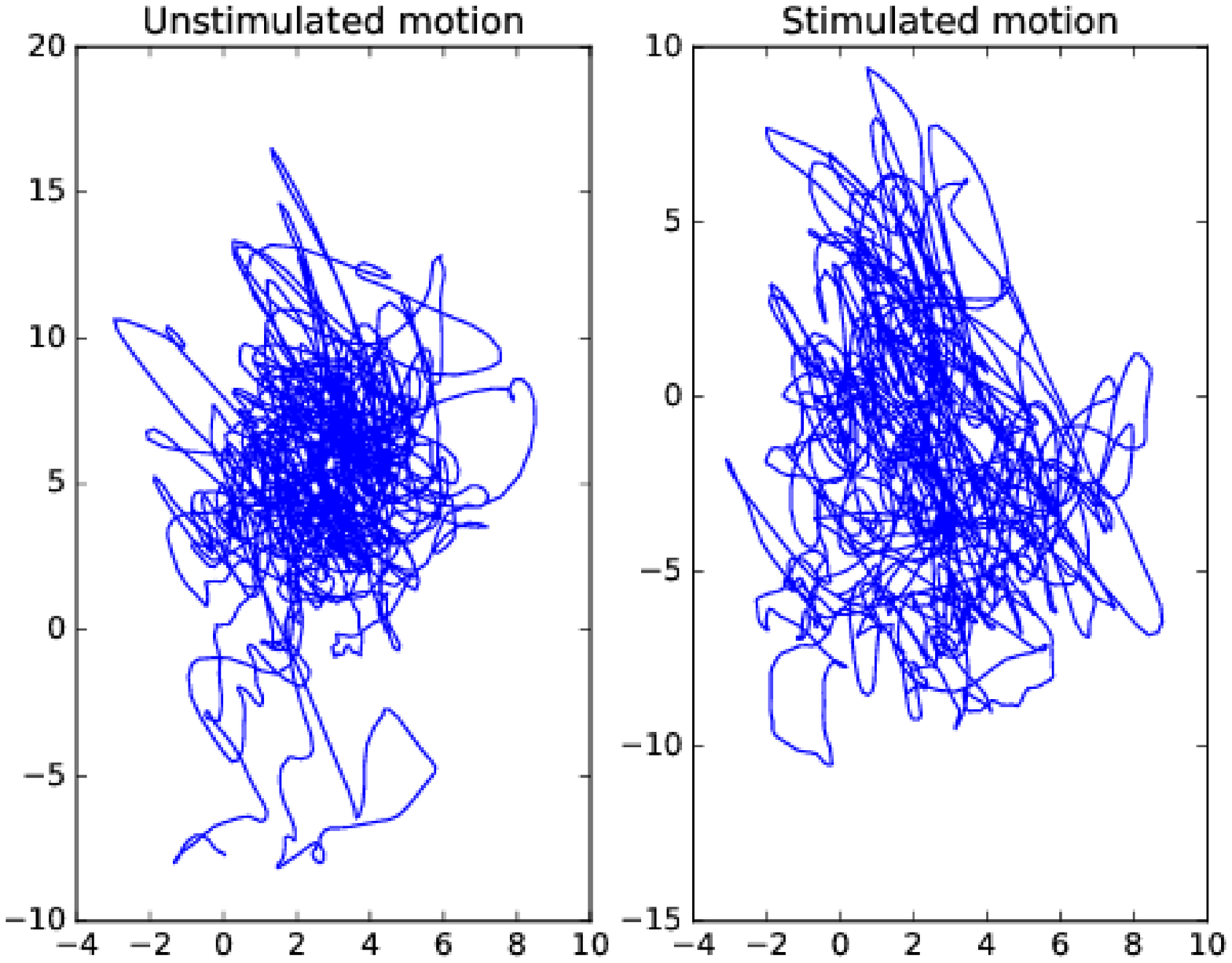}
  \caption{Patient 6 under eight directional instructions: "unstimulated motion" means that the
  patient moves without transcutaneous stimulations, while "stimulated motion" represents the motion
  under stimulations.}
  \label{fig:patient6}
  \vspace{-0.5cm}

\end{figure}

The COP sensory data from each session is discretized on a $100\times 100$ grid, which is fine
enough to capture the patient's small movements. The problem is formulated into a MDP, where each
state captures the patient's discretized location and velocity, and the set of actions changes the
velocity into eight possible directions. The velocity is represented with a two-dimensional vector
showing eight possible velocity directions. Thus the problem has 80000 states and 8 actions, and
each action is assumed to lead to a deterministic state.

\begin{table*}
  \centering
  \caption{Evaluation of the learned rewards: "forward" etc. denote the instructed direction; item
  name"1u" denotes the patient id "1", with "u" denoting unstimulated session and "s" denoting
  stimulated sessions. The table shows the correlation coefficient between the ideal reward and the
  recovered reward.}
  \label{tab:feature1}
 \begin{tabular}{|c|c|c|c|c|c|c|c|c|c|}
      \hline
     &forward&backward&left&right&top left&top right&bottom left&bottom right&origin\\\hline
   1u&0.411741&0.257564&0.0691989&-0.210216&&&&&0.49016\\\hline
   1s&0.200355&0.486723&0.129839&0.436533&&&&&0.207188\\\hline
   2u&0.161595&-0.17814&0.153376&-0.16767&0.162906&0.105993&-0.0211192&-0.220457&0.156034\\\hline
   2s&-0.0310265&-0.0803484&0.0474505&-0.00146299&0.0442916&0.0874981&0.00668849&0.0742221&0.0726437\\\hline
   3u&0.362801&-0.2995&0.245916&-0.178778&0.386421&0.0148849&-0.00335653&-0.385605&0.0507719\\\hline
   3s&-0.265834&0.146516&0.379665&-0.272437&0.138805&-0.2683&0.212331&0.00301386&-0.182916\\\hline
   4u&0.301472&-0.281474&0.377787&-0.320403&0.410212&-0.119599&0.136309&-0.306677&0.171433\\\hline
   4s&-0.104719&0.0930068&0.327783&-0.229091&0.175432&-0.161819&0.323862&-0.0521654&-0.202197\\\hline
   5u&0.360293&-0.311692&-0.253715&0.260426&0.0863029&0.495134&-0.38137&-0.140836&-0.160687\\\hline
   5s&-0.212823&0.0414435&0.0908994&-0.124174&0.00414109&-0.107462&0.122018&0.0453461&0.145686\\\hline
   6u&-0.0416432&0.0570847&0.210028&-0.104113&0.0363181&-0.0672399&0.0704143&-0.00392284&0.190253\\\hline
   6s&-0.157148&0.178879&0.0880393&-0.0718817&-0.102579&-0.298918&0.307328&0.171319&0.359168\\\hline
 \end{tabular}
\end{table*}

To learn the reward function from the observed trajectories based on the formulated MDP, we use the
coordinate and velocity direction of each grid as the feature, and learn the reward function
parameter from each set of data after segmentation based on peak detection on distances from the
origin. The function approximator is a neural network with three hidden layers and $[100,50,25]$
nodes. The huber loss function is used in reward sharing, and the result is show in Table
\ref{tab:feature1}.

It shows that the patient's ability to following instructions vary among different directions, and
the values will assist physicians to design the stimulating signals.
}
\section{Conclusions}
\label{irl::conclusions}
This work proposes a solution to learn an accurate reward function for each task with limited
demonstrations but from the same demonstrator, by maximizing the shared rewards among different
tasks. We proposed several loss functions to maximize the shared reward, and compared their
accuracies in a simulated environment. It shows that huber loss has the best
performance.

In future work, we will apply the proposed method to imitation learning.


\begin{thebibliography}{10}
\providecommand{\url}[1]{#1}
\csname url@samestyle\endcsname
\providecommand{\newblock}{\relax}
\providecommand{\bibinfo}[2]{#2}
\providecommand{\BIBentrySTDinterwordspacing}{\spaceskip=0pt\relax}
\providecommand{\BIBentryALTinterwordstretchfactor}{4}
\providecommand{\BIBentryALTinterwordspacing}{\spaceskip=\fontdimen2\font plus
\BIBentryALTinterwordstretchfactor\fontdimen3\font minus
  \fontdimen4\font\relax}
\providecommand{\BIBforeignlanguage}[2]{{%
\expandafter\ifx\csname l@#1\endcsname\relax
\typeout{** WARNING: IEEEtran.bst: No hyphenation pattern has been}%
\typeout{** loaded for the language `#1'. Using the pattern for}%
\typeout{** the default language instead.}%
\else
\language=\csname l@#1\endcsname
\fi
#2}}
\providecommand{\BIBdecl}{\relax}
\BIBdecl

\bibitem{irl::irl1}
A.~Y. Ng and S.~Russell, ``Algorithms for inverse reinforcement learning,'' in
  \emph{in Proc. 17th International Conf. on Machine Learning}, 2000.

\bibitem{irl::rl}
R.~S. Sutton and A.~G. Barto, \emph{Reinforcement learning: An
  introduction}.\hskip 1em plus 0.5em minus 0.4em\relax MIT press Cambridge,
  1998, vol.~1, no.~1.

\bibitem{irl::irl2}
P.~Abbeel and A.~Y. Ng, ``Apprenticeship learning via inverse reinforcement
  learning,'' in \emph{Proceedings of the twenty-first international conference
  on Machine learning}.\hskip 1em plus 0.5em minus 0.4em\relax ACM, 2004, p.~1.

\bibitem{irl::motionanalysis}
B.~Najafi, K.~Aminian, A.~Paraschiv-Ionescu, F.~Loew, C.~J. Bula, and
  P.~Robert, ``Ambulatory system for human motion analysis using a kinematic
  sensor: monitoring of daily physical activity in the elderly,'' \emph{IEEE
  Transactions on biomedical Engineering}, vol.~50, no.~6, pp. 711--723, 2003.

\bibitem{irl::kalman}
R.~Kalman and M.~M. C. B. D. R.~I. for Advanced Studies. Center~for
  Control~Theory, \emph{When is a Linear Control System Optimal?.}, ser. RIAS
  technical report.\hskip 1em plus 0.5em minus 0.4em\relax Martin Marietta
  Corporation, Research Institute for Advanced Studies, Center for Control
  Theory, 1963.

\bibitem{irl::maxmargin}
N.~D. Ratliff, J.~A. Bagnell, and M.~A. Zinkevich, ``Maximum margin planning,''
  in \emph{Proceedings of the 23rd international conference on Machine
  learning}.\hskip 1em plus 0.5em minus 0.4em\relax ACM, 2006, pp. 729--736.

\bibitem{irl::maxentropy}
B.~D. Ziebart, A.~Maas, J.~A. Bagnell, and A.~K. Dey, ``Maximum entropy inverse
  reinforcement learning,'' in \emph{Proc. AAAI}, 2008, pp. 1433--1438.

\bibitem{irl::sequence}
Q.~P. Nguyen, B.~K.~H. Low, and P.~Jaillet, ``Inverse reinforcement learning
  with locally consistent reward functions,'' in \emph{Advances in Neural
  Information Processing Systems}, 2015, pp. 1747--1755.

\bibitem{irl::gaussianirl}
S.~Levine, Z.~Popovic, and V.~Koltun, ``Nonlinear inverse reinforcement
  learning with gaussian processes,'' in \emph{Advances in Neural Information
  Processing Systems 24}, J.~Shawe-Taylor, R.~S. Zemel, P.~L. Bartlett,
  F.~Pereira, and K.~Q. Weinberger, Eds.\hskip 1em plus 0.5em minus 0.4em\relax
  Curran Associates, Inc., 2011, pp. 19--27.

\bibitem{irl::guidedirl}
C.~Finn, S.~Levine, and P.~Abbeel, ``Guided cost learning: Deep inverse optimal
  control via policy optimization,'' \emph{arXiv preprint arXiv:1603.00448},
  2016.

\bibitem{irl::pomdp}
J.~Choi and K.-E. Kim, ``Inverse reinforcement learning in partially observable
  environments,'' \emph{Journal of Machine Learning Research}, vol.~12, no.
  Mar, pp. 691--730, 2011.

\bibitem{irl::localirl}
S.~Levine and V.~Koltun, ``Continuous inverse optimal control with locally
  optimal examples,'' \emph{arXiv preprint arXiv:1206.4617}, 2012.

\bibitem{irl::deepirl}
M.~Wulfmeier, P.~Ondruska, and I.~Posner, ``Deep inverse reinforcement
  learning,'' \emph{arXiv preprint arXiv:1507.04888}, 2015.

\bibitem{irl::bayirl}
D.~Ramachandran and E.~Amir, ``Bayesian inverse reinforcement learning,'' in
  \emph{Proceedings of the 20th International Joint Conference on Artifical
  Intelligence}, ser. IJCAI'07.\hskip 1em plus 0.5em minus 0.4em\relax San
  Francisco, CA, USA: Morgan Kaufmann Publishers Inc., 2007, pp. 2586--2591.

\bibitem{irl::subgradient}
G.~Neu and C.~Szepesv{\'a}ri, ``Apprenticeship learning using inverse
  reinforcement learning and gradient methods,'' \emph{arXiv preprint
  arXiv:1206.5264}, 2012.

\bibitem{irl::bayioc}
K.~Mombaur, A.~Truong, and J.-P. Laumond, ``From human to humanoid
  locomotion—an inverse optimal control approach,'' \emph{Autonomous robots},
  vol.~28, no.~3, pp. 369--383, 2010.

\bibitem{irl::multirl1}
C.~Dimitrakakis and C.~A. Rothkopf, ``Bayesian multitask inverse reinforcement
  learning,'' in \emph{European Workshop on Reinforcement Learning}.\hskip 1em
  plus 0.5em minus 0.4em\relax Springer, 2011, pp. 273--284.

\bibitem{irl::multirl2}
J.~Choi and K.-E. Kim, ``Nonparametric bayesian inverse reinforcement learning
  for multiple reward functions,'' in \emph{Advances in Neural Information
  Processing Systems}, 2012, pp. 305--313.

\bibitem{irl::relative}
A.~Boularias, J.~Kober, and J.~R. Peters, ``Relative entropy inverse
  reinforcement learning,'' in \emph{International Conference on Artificial
  Intelligence and Statistics}, 2011, pp. 182--189.

\bibitem{irl::fairl}
K.~Li and J.~W. Burdick, ``Large-scale inverse reinforcement learning via
  function approximation for clinical motion analysis,'' \emph{arXiv preprint
  arXiv:1707.09394}, 2017.

\bibitem{irl::metasurveyclassification}
R.~Vilalta and Y.~Drissi, ``A perspective view and survey of meta-learning,''
  \emph{Artificial Intelligence Review}, vol.~18, no.~2, pp. 77--95, 2002.

\bibitem{irl::metamemory}
A.~Santoro, S.~Bartunov, M.~Botvinick, D.~Wierstra, and T.~Lillicrap,
  ``Meta-learning with memory-augmented neural networks,'' in
  \emph{International conference on machine learning}, 2016, pp. 1842--1850.

\bibitem{irl::representation}
B.~Hariharan and R.~Girshick, ``Low-shot visual recognition by shrinking and
  hallucinating features,'' \emph{arXiv preprint arXiv:1606.02819}, 2016.

\bibitem{irl::learngradient}
M.~Andrychowicz, M.~Denil, S.~Gomez, M.~W. Hoffman, D.~Pfau, T.~Schaul, and
  N.~de~Freitas, ``Learning to learn by gradient descent by gradient descent,''
  in \emph{Advances in Neural Information Processing Systems}, 2016, pp.
  3981--3989.

\bibitem{irl::hypernetworks}
D.~Ha, A.~Dai, and Q.~V. Le, ``Hypernetworks,'' \emph{arXiv preprint
  arXiv:1609.09106}, 2016.

\bibitem{irl::metarl}
N.~Schweighofer and K.~Doya, ``Meta-learning in reinforcement learning,''
  \emph{Neural Networks}, vol.~16, no.~1, pp. 5--9, 2003.

\bibitem{irl::metamultirl}
E.~Parisotto, J.~L. Ba, and R.~Salakhutdinov, ``Actor-mimic: Deep multitask and
  transfer reinforcement learning,'' \emph{arXiv preprint arXiv:1511.06342},
  2015.

\bibitem{irl::rl2}
Y.~Duan, J.~Schulman, X.~Chen, P.~L. Bartlett, I.~Sutskever, and P.~Abbeel,
  ``Rl2: Fast reinforcement learning via slow reinforcement learning,''
  \emph{arXiv preprint arXiv:1611.02779}, 2016.

\bibitem{irl::maml}
C.~Finn, P.~Abbeel, and S.~Levine, ``Model-agnostic meta-learning for fast
  adaptation of deep networks,'' \emph{arXiv preprint arXiv:1703.03400}, 2017.

\bibitem{irl::osil}
Y.~Duan, M.~Andrychowicz, B.~Stadie, J.~Ho, J.~Schneider, I.~Sutskever,
  P.~Abbeel, and W.~Zaremba, ``One-shot imitation learning,'' \emph{arXiv
  preprint arXiv:1703.07326}, 2017.

\bibitem{irl::BGI}
K.~{Li} and J.~W. {Burdick}, ``{Bellman Gradient Iteration for Inverse
  Reinforcement Learning},'' \emph{ArXiv e-prints}, Jul. 2017.

\bibitem{irl::spi}
S.~Harkema, Y.~Gerasimenko, J.~Hodes, J.~Burdick, C.~Angeli, Y.~Chen,
  C.~Ferreira, A.~Willhite, E.~Rejc, R.~G. Grossman \emph{et~al.}, ``Effect of
  epidural stimulation of the lumbosacral spinal cord on voluntary movement,
  standing, and assisted stepping after motor complete paraplegia: a case
  study,'' \emph{The Lancet}, vol. 377, no. 9781, pp. 1938--1947, 2011.

\end{thebibliography}
\end{document}